\pdfoutput=1
\documentclass[11pt]{article}

\usepackage{acl}
\usepackage{tabularx}
\usepackage{times}
\usepackage{latexsym}
\usepackage{comment}
\usepackage{booktabs}
\usepackage{graphicx}
\usepackage{refcount}
\usepackage{amsmath}
\usepackage{dsfont}
\usepackage{hyperref}

\usepackage[T1]{fontenc}
\usepackage[utf8]{inputenc}

\usepackage{microtype}
\usepackage{multirow}
\usepackage{xcolor}

\definecolor{myred}{RGB}{233, 69, 96}
\definecolor{myorange}{RGB}{242, 161, 84}
\definecolor{myblue}{RGB}{89, 111, 183}
\definecolor{mygreen}{RGB}{46, 176, 134}
\title{Extroversion or Introversion? Controlling The Personality of Your Large Language Models}

\author{Yanquan Chen, Zhen Wu\thanks{\ \ Corresponding author.}, Junjie Guo, Shujian Huang, Xinyu Dai \\
  National Key Laboratory for Novel Software Technology, Nanjing University, China \\
  School of Artificial Intelligence, Nanjing University, China\\
  \texttt{\{yanquan.chen, guojj\}@smail.nju.edu.cn} \\
  \texttt{\{wuz, huangsj, daixinyu\}@nju.edu.cn} \\
}

\begin{document}
\maketitle
\begin{abstract}

Large language models (LLMs) exhibit robust capabilities in text generation and comprehension, mimicking human behavior and exhibiting synthetic personalities. However, some LLMs have displayed offensive personality, propagating toxic discourse.
Existing literature neglects the origin and evolution of LLM personalities, as well as the effective personality control. 
To fill these gaps, our study embarked on a comprehensive investigation into LLM personality control. 
We investigated several typical methods to influence LLMs, including three training methods: Continual Pre-training, Supervised Fine-Tuning (SFT), and Reinforcement Learning from Human Feedback (RLHF), along with inference phase considerations (prompts).
Our investigation revealed a hierarchy of effectiveness in control: Prompt > SFT > RLHF > Continual Pre-train.
Notably, SFT exhibits a higher control success rate compared to prompt induction.
While prompts prove highly effective, we found that prompt-induced personalities are less robust than those trained, making them more prone to showing conflicting personalities under reverse personality prompt induction.
Besides, harnessing the strengths of both SFT and prompt, we proposed \underline{P}rompt \underline{I}nduction post \underline{S}upervised \underline{F}ine-tuning (PISF), which emerges as the most effective and robust strategy for controlling LLMs' personality, displaying high efficacy, high success rates, and high robustness. 
Even under reverse personality prompt induction, LLMs controlled by PISF still exhibit stable and robust personalities. 
Codes and datasets are available at \href{https://github.com/DespairL/Personality}{here}.

\end{abstract}

\section{Introduction}

With the rapid advancement of large-scale pre-training~~\cite{kaplanScalingLawsNeural2020,brownLanguageModelsAre2020,chowdheryPaLMScalingLanguage2022}, large language models 
\begin{figure}[htbp]
  \centering
  \includegraphics[width=0.48\textwidth]{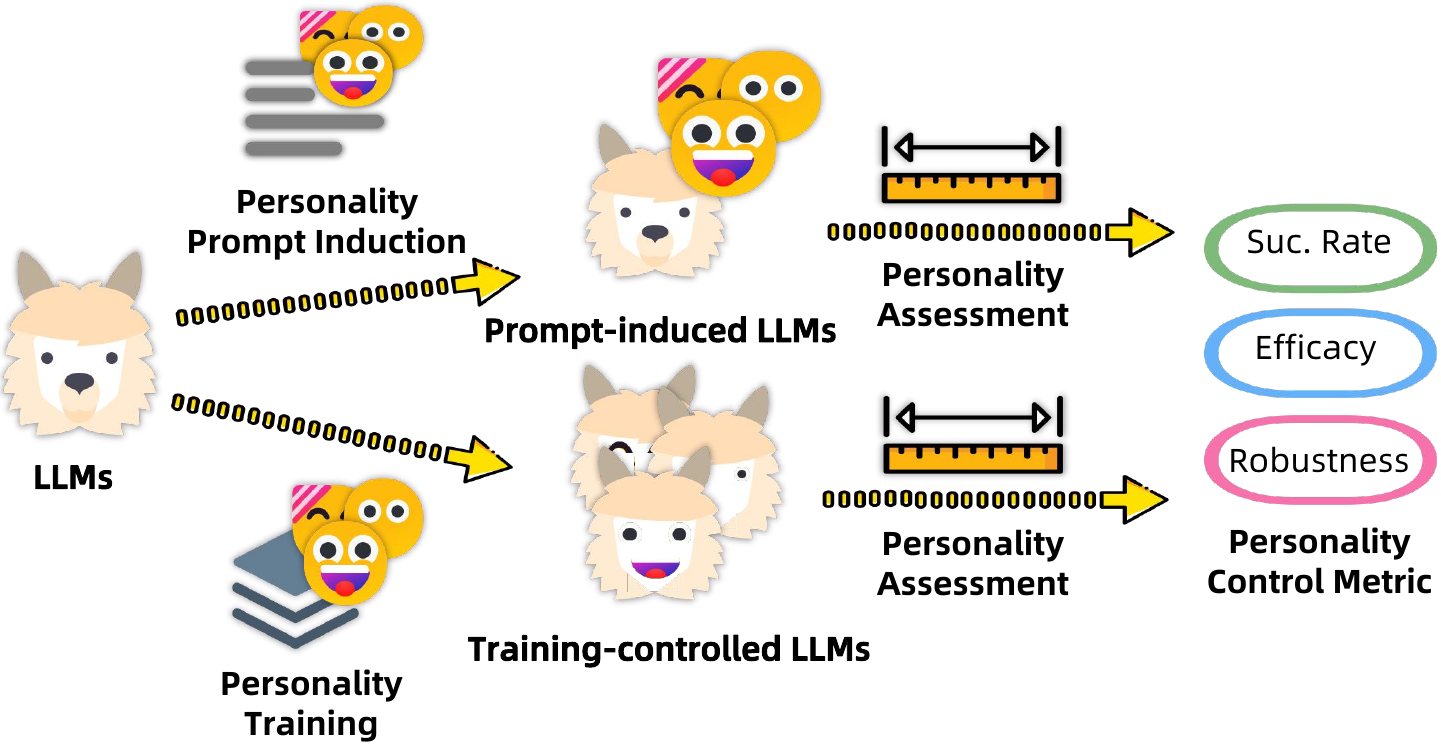}
  \caption{Overview. We embarked on a comprehensive investigation into personality control with typical methods to influence LLMs.}
  \label{fig:overview}
\end{figure}
(LLMs) have made significant strides in natural language processing, demonstrating robust capabilities in text generation and 
comprehension~\cite{wei2022emergent}. 
Enabled by vast amounts of training data, LLMs mimic human characteristics in outputs, showcasing synthetic personalities~\cite{serapio-garciaPersonalityTraitsLarge2023a}.
However, variations in architecture, training data, and methodologies yield distinct synthetic personalities among different LLMs~\cite{miotto-etal-2022-gpt, pan2023llms}. Despite their capabilities, some LLMs have displayed offensive personality, propagating toxic discourse~\cite{wen2023unveiling, 10.1145/3531146.3533229, deshpande-etal-2023-toxicity}. These concerns surrounding LLMs' synthetic personalities have garnered widespread attention in AI safety and psychology research~\cite{hagendorff2023machine, demszky_using_2023}.

%\footnotetext{Preprint. Code and data will be available soon.}

Currently, the mechanisms underlying LLM personality formation remain largely unexplored. 
Previous community efforts have primarily focused on validating human personality assessments on LLMs, supported by psychological theories~\cite{serapio-garciaPersonalityTraitsLarge2023a, huang2023revisiting}, adapting human personality assessment to characterize LLM personalities~\cite{miotto-etal-2022-gpt, pan2023llms} and exploring personality assessments suitable for LLMs~\cite{jiang2023evaluating}. 
% personality measurements in the outputs of some LLMs under specific prompting configurations are reliable and valid
Notably, \citeauthor{serapio-garciaPersonalityTraitsLarge2023a}(\citeyear{serapio-garciaPersonalityTraitsLarge2023a}) found that 
personality assessments in the outputs of some LLMs are reliable and valid.
%human psychological assessments can evaluate the LLMs' synthetic personality, aligning with the standards of psychological evaluations.
Additionally, a few studies have explored inducing personality in LLMs through prompts or fine-tuning~\cite{karra2023estimating, serapio-garciaPersonalityTraitsLarge2023a, huang2023revisiting}.

\begin{figure*}[htbp]
  \centering
  \includegraphics[width=0.75\textwidth]{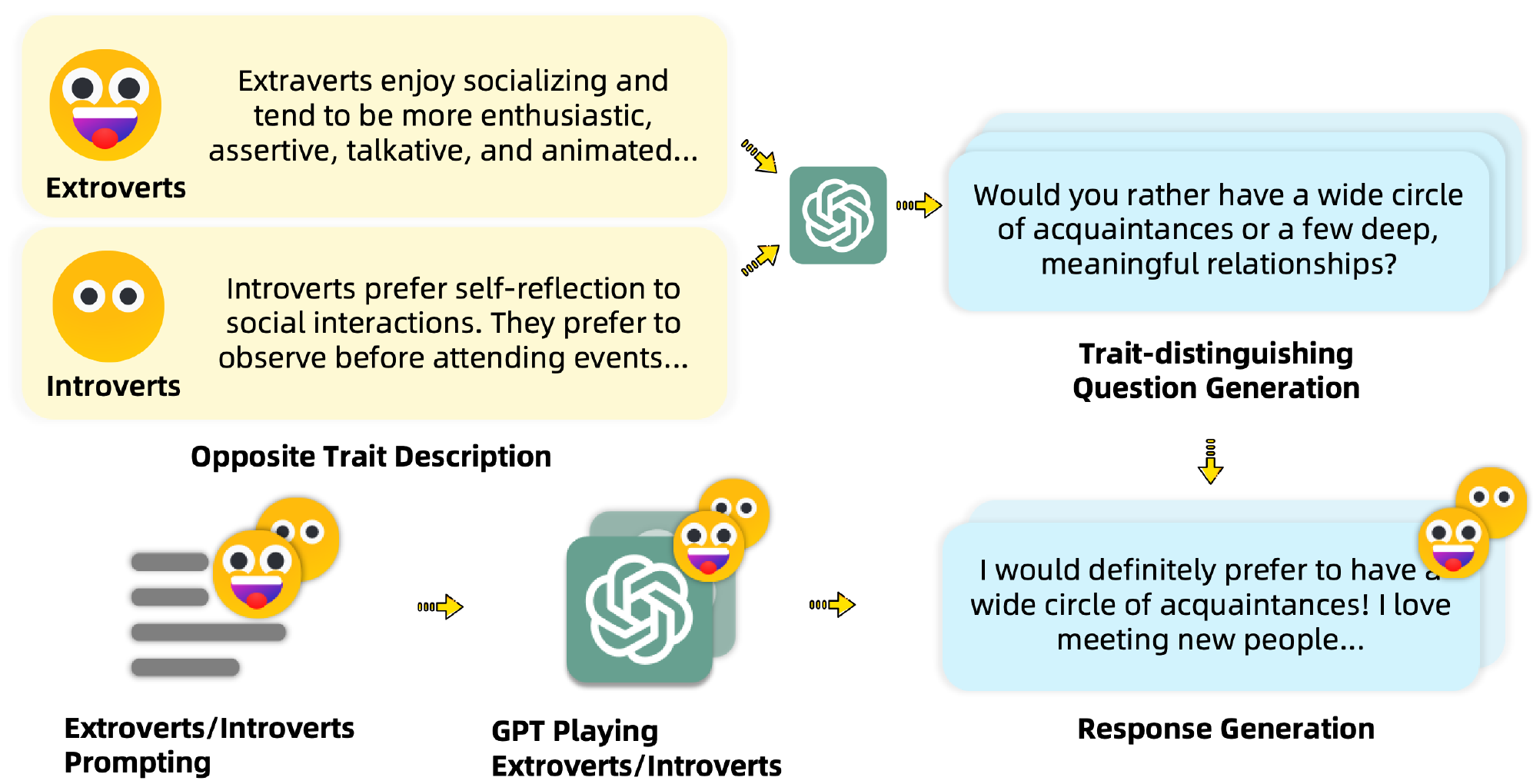}
  \caption{Instruction Data Generation with Prompt-induced LLMs. Utilizing the Least-to-Most~\cite{zhou2023leasttomost} mindset, we partitioned the data generation process into two stages: initially crafting questions rooted in Opposite Trait Description, followed by eliciting responses from Prompt-induced LLMs.}
  \label{fig:data_generation_pipeline}
\end{figure*}

%However, existing literature neglects the origin and evolution of LLMs' synthetic personalities, as well as the effective personality control. 
However, existing literature neglects how to effectively control the personality of LLMs and ensure its stability and resistance to alteration.
Filling these gaps is crucial due to the immense potential to utilize LLMs with well-defined and consistent personalities. This enables customization of LLMs' synthetic personalities to suit specific contextual requirements.
For example, rational LLMs might excel in logical reasoning tasks, while empathetic LLMs could be ideal for companion robots. Based on these considerations, in this work, we explore two questions: 1) \textit{During building and using LLMs, what factor has a greater impact on shaping LLMs' synthetic personality?} 2) \textit{How to control LLMs' synthetic personality effectively and robustly?}

%The importance of addressing these inquiries is underscored by the vast potential to deploy LLMs with defined and consistent personalities tailored to specific contextual needs (e.g., rational LLMs for logical reasoning, or enthusiastic and empathetic LLMs for companion robots ...).

To answer these questions, we examined synthesized personality control using several typical factors, encompassing three training methods (Continual Pre-train~\cite{HAN2021225}, Supervised Fine-Tuning (SFT), Reinforcement Learning from Human Feedback (RLHF)~\cite{ouyang2022training, bai2022training}), along with inference phase considerations (prompts) -- guided by MBTI theory~\cite{myersMyersBriggsTypeIndicator1962,e73f6545-174d-366a-9e0e-d380358fcd01, articleMcCraeReinterpreting}.  
The MBTI theory categorizes individuals into sixteen \textbf{personality types}, delineated by preferences across four dichotomous dimensions.
Each dichotomous dimension comprises two opposite \textbf{personality traits}.
For instance, the Attitude dimension comprises two traits -- \textit{Extroversion} vs. \textit{Introversion}.
The trait combinations on four dimensions determine a specific personality type (e.g., ENFP personality type: \textit{Extraversion}, \textit{Intuition}, \textit{Feeling}, \textit{Perceiving}).
Thus, we can naturally differentiate control targets into overall personality types or local traits. 
In this work, we delved into \textbf{Specific Trait Control} and \textbf{Specific Personality Control}.

To study LLM synthesized personality control, we constructed corresponding trait datasets and personality datasets, e.g. 2500 trait instructions for SFT and 20,000 paired trait data for the RLHF Reward Model, which can significantly serve our research. To measure changes in personality pre- and post-control of LLMs, we specifically designed quantitative metrics, involving Induction Success Rate ($\mathrm{ISR}$) and Trait Induction Efficacy ($\mathrm{TIE}$) for assessing the effectiveness of trait control, Trait Stabilization Efficacy ($\mathrm{TSE}$) to monitor fluctuations induced by trait control, Personality Induction Success Rate ($\mathrm{PISR}$) and Personality Induction Efficacy ($\mathrm{PIE}$) for assessing the effectiveness of personality control. We visually illustrate our main exploration in Figure \ref{fig:overview}.

Our investigation unveiled a hierarchy of effectiveness in LLM personality control: Prompt > SFT > RLHF > Continual Pre-train. 
Particularly, SFT demonstrates a higher control success rate compared to prompt induction. 
While prompts prove highly effective, we note that prompt-induced personality is less robust than those shaped through training, rendering them susceptible to conflicting personality shifts via reverse personality prompt induction. Besides, by leveraging the strengths of both SFT and prompts, we proposed \underline{P}rompt \underline{I}nduction post \underline{S}upervised \underline{F}ine-tuning (PISF), which enhances control effectiveness and personality robustness, characterized by high efficacy, high success rates, and high robustness. 
Even when exposed to reverse personality prompt induction, personalities under PISF control demonstrate resistance to change.

% RPPI要不要算做contribution
Our contributions can be summarized as follows:
\begin{itemize}
    \item We are the first to systematically investigate the factors influencing LLM personalities and effective methods for controlling them.
    \item Our investigation unveiled a hierarchy of effectiveness in control: Prompt > SFT > RLHF > Continual Pre-train. Additionally, we proposed \underline{P}rompt \underline{I}nduction post \underline{S}upervised \underline{F}ine-tuning (PISF), which emerges as the most effective and robust method for controlling synthetic personalities and exhibits high efficacy, high success rates, and high robustness.
    \item We provided a comprehensive dataset containing all traits and personalities, facilitating thorough exploration of each training method. 
    We proposed several quantitative metrics to evaluate the effectiveness of specific trait control and specific personality control. These contributions will accelerate research in the field.
\end{itemize}

\section{Background: Personality Assessment}

In this section, we introduce two key personality models widely used in research: the Myers-Briggs Type Indicator~\cite{myersMyersBriggsTypeIndicator1962,e73f6545-174d-366a-9e0e-d380358fcd01, articleMcCraeReinterpreting} and the Big Five~\cite{goldbergAlternativeDescriptionPersonality1990}.
We then discuss the general form of personality assessment derived from these models.
Lastly, we provide a concise overview of how personality assessment is approached in our study.

% Personality Models -> Personality Assessment Scales -> Our Work
\noindent \textbf{Big Five} The Big Five model emerged from lexical analysis of English personality adjectives~\cite{goldbergAlternativeDescriptionPersonality1990}. It encompasses five key dimensions: Openness (O), Conscientiousness (C), Extraversion (E), Agreeableness (A), and Neuroticism (N). Big Five assessments yield continuous and quantitative scores for each factor (e.g. O: $47\%$, C: $64\%$, E: $51\%$, A: $52\%$, N: $38\%$), posing difficulties in categorizing and studying similar individual profiles.

\noindent \textbf{Myers-Briggs Type Indicator} The MBTI theory, stemming from Carl Jung's seminal work~\cite{jungPsychologicalTypes1923}, categorizes individuals into sixteen personality types based on preferences across four dichotomous dimensions (Extroversion/Introversion, Sensing/Intuition, Thinking/Feeling, and Judging/Perceiving). Each individual's personality type is characterized by a profile of four traits (e.g., ENFP: Extraversion, Intuition, Feeling, Perceiving) representing their preferences in each dimension. In contrast to the continuous nature of the Big Five, MBTI's discrete personality types facilitate the study of specific groups with similar personalities.

\noindent \textbf{General Form of Personality Assessment} Personality assessments commonly consist of Likert Items~\cite{likertTechniqueMeasurementAttitudes1932} crafted according to established personality models. Likert Items are statements or questions presented to respondents for evaluation, typically utilizing a five to seven-point scale to gauge agreement or disagreement ~\cite{kulasMiddleResponseFunctioning2008}. The form is known as Likert Scale. As shown in Table \ref{evaluate_prompt}, \textit{People who know you tend to describe you as} and the following options represent a Likert Item. And \textcolor{myblue}{Task Description} delineates different levels of agreement. We can map the level of agreement in responses following the \textcolor{myorange}{Item Postamble} to 5-point scale.

\noindent \textbf{Personality Assessment in Our Work} For research convenience, we utilized the MBTI personality model for assessments and dataset construction. To ensure reliability, we collected publicly available questionnaires~\cite{pan2023llms} and refined them into a 200-item MBTI Assessment (50 items per dichotomous dimension). We detailed the format and sources of the questionnaires in the Appendix \ref{sec:mbti_personality_assessment} for further reference. The rating scale was designed based on 5-point Likert Scale. (\S \ref{sec:evaluation_details}) 

\section{Methodology}

To delve into the control of LLMs' synthesized personalities during training, we constructed datasets for three training methods based upon MBTI~\cite{myersMyersBriggsTypeIndicator1962}(\S \ref{training_data}). For research convenience, we utilized the MBTI personality model for assessments (\S \ref{sec:evaluation_details}). To measure personality variations pre- and post-controlling LLMs, we proposed several quantitative metrics for specific trait control and specific personality control (\S \ref{sec:indicators}). 
% 移除训练这段直接移到setting里面，并且前面就讲宏观层面的训练目标
% Leveraging these datasets, personality assessments and quantitative metrics, we conducted exploration across the aforementioned stages(\S \ref{training_method}).

\subsection{Personality Dataset Construction for Popular Training Methods}\label{training_data}

We utilized popular training methods at each stage - autoregressive for continual pre-train, instruction tuning for SFT and PPO for RLHF. To meet the requirements of various training methods, we constructed corresponding personality datasets. 

\noindent \textbf{Continual Pre-training.} We continual pretrained LLMs using widely adopted autoregressive objective~\cite{radford2019language, brownLanguageModelsAre2020}. To elaborate, the model is trained to predict the next token in a sequence of continuous text by leveraging the surrounding context.

\begin{figure}[htbp]
  \centering
  \includegraphics[width=0.48\textwidth]{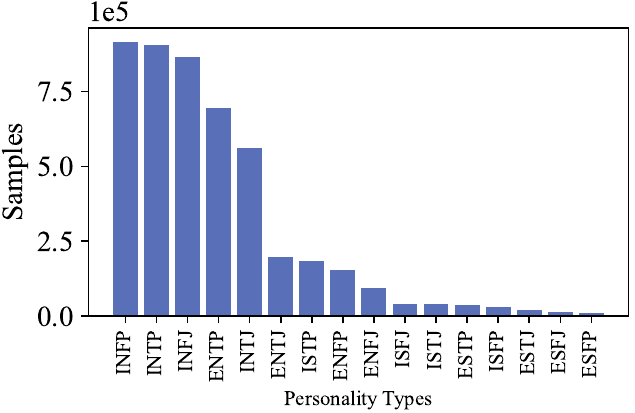}
  \caption{Pretrain Data Distribution.}
  \label{fig:pretrain_data}
\end{figure}

Due to the difficulty of generating a large amount of long-context pre-training data, we amalgamated and refined existing datasets annotated with human personality~\cite{dylan_storey_2018_1482951}. 
The data distribution can be referred to Figure \ref{fig:pretrain_data}. 
%Please refer to Appendix \ref{sec:training_data} for more details.

% 要不要缩写 specific personality control / specific trait control
Intuitively, a control method that achieves better performance with less data is preferable. Thus, we investigated the control of continual pre-training with limited data. We randomly sampled 10,000 instances from human labeled data for each personality and aggregated data of eight personalities with the target trait as the trait data. Specifically, we aggregated 10,000 ENFJ, ENFP, ENTJ, ENTP, ESFJ, ESFP, ESTJ, ESTP personality data as E trait data. Thus, each trait comprises 80,000 instances.

\noindent \textbf{SFT.} We adopted widely used instruction tuning~\cite{wei2022finetuned, alpaca, vicuna2023} as the training objective. Instruction tuning refers to the process of further training LLMs on a dataset consisting of (instruction, output) pairs in a supervised fashion, which bridges the gap between the next-word prediction objective and the objective of making LLMs adhere to human instruction~\cite{zhang2024instruction}. 

Akin to numerous studies utilizing LLMs for data generation~\cite{wang-etal-2023-self-instruct, alpaca, lee-etal-2023-ensemble}, as shown in Figure \ref{fig:data_generation_pipeline}, we utilized prompt-induced LLMs to generate training data. 
% 增加数据生成过程的描述
To enhance data quality, inspired by Least-to-Most~\cite{zhou2023leasttomost}, we divided the data generation process into two stages: 
In Stage 1, to assist the model in distinguishing between opposite traits, we incorporated descriptions of two opposite traits belonging to the same dimension into the prompt, followed by prompting the model to generate questions. 
In Stage 2, we employed prompt-induced models to respond to respective questions. 
Then we integrated the obtained questions and responses as SFT dataset. To ascertain the capacity of LLMs in producing data with specific personalities, we conducted a preliminary investigation (\S \ref{sec:preliminary}) to validate the prompt induction proficiency based on personality description.  
% 附录内容后面再看
%We presented generation prompt examples in the Appendix \ref{sec:template_data_generation} and instruction pair example in Appendix~\ref{sec:training_data}. 

For SFT, the quantity of data is not critical~\cite{zhou2023lima}. Thus, we utilized prompt-induced GPT-3.5-turbo-1106 to generate 2,500 instances for each trait and aggregated four trait data for each personality type. Specifically, we aggregated instances of E, N, T and J as ENTJ personality data. Therefore, each personality has 10,000 instances. 

% 增加RLHF数据生成的细节
\noindent \textbf{RLHF.} Following previous work~\cite{ziegler2020finetuning, ouyang2022training}, we initially trained reward model directly from feedback, and subsequently utilizes it as a reward function to enhance an agent's policy via widely-used proximal policy optimization (PPO). 
In our work, the reward model is trained in a supervised manner to classify responses to prompts as either conforming to the target trait or personality (high reward) or not (low reward). 
We constructed datasets for PPO training and reward training.

For PPO, to control variables, we used the same questions from SFT data as input. 
And for reward model, we used prompt-induced LLMs to generate question-answer pairs. We instructed models to provide responses to the same question but with opposing traits, establishing a ranking relationship. For instance, we trained extroversion reward model with (Question, Extroversion-induced model response, Introversion-induced model response). 

Inspired by InstructGPT~\cite{ouyang2022training}, we generated in-distribution data to fit the model distribution as well as out-of-distribution data for generalization.
Specifically, for each trait, we used prompt-induced GPT-3.5-turbo-1106 to generate 5000 pairs as out-of-distribution data and employed prompt-induced Llama2-chat-13B and ChatGLM2-6B to generate 15,000 pairs as in-distribution data. 
The total comprises 20,000 pairs for each trait. Similar to SFT, we integrate trait data to obtain personality data, i.e. 80,000 pairs for each personality.

% 下面这段移到训练那里，这里只讲人格数据
%And previous work has shown that reinforcement learning training process may cause models to suffer from performance decrease in language ability~\cite{ouyang2022training}. Therefore, we followed previous work~\cite{yao2023dschat, ouyang2022training} by incorporating autoregressive training into the reinforcement learning process with widely recognized Wikipedia datasets~\cite{wikidump}, ensuring the model retains its ability to generate responses smoothly. For Wikipedia datasets, we provide more details in the Appendix~\ref{sec:training_data}. 

%For the reward model, inspired by InstructGPT~\cite{ouyang2022training}, we used prompt-induced GPT-3.5-turbo-1106 to generate 5000 data pairs as out-of-distribution data for generalization.  Additionally, prompt-induced Llama2-chat-13B and ChatGLM2-6B were employed to generate 15,000 data pairs to fit the model distribution. Given that Actor and Reward models share embedding space, we trained the reward model for each corresponding model. We presented reward data examples in the Appendix~\ref{sec:training_data} and the reward model performance in the Appendix \ref{sec:rlhf_training}.  

\begin{table}[htbp]
    \centering
    % \resizebox{0.5\textwidth}{!}{
    \small
    \begin{tabular}{ccccc}
    \toprule
    \multirow{2}{*}{\textbf{Dataset}} & \multicolumn{2}{c}{\textbf{Trait}} & \multicolumn{2}{c}{\textbf{Personality}}\\
    \cmidrule(r){2-3} \cmidrule(l){4-5} 
    & \textbf{Train} & \textbf{Valid} & \textbf{Train} & \textbf{Valid} \\
    \midrule
    Pre-train & 80000 & - & 10000 & - \\
    SFT & 2500 & - & 10000 & - \\
    RLHF-PPO & 2500 & - & 10000 & - \\
    RLHF-Reward & 18000 & 2000 & 72000 & 8000 \\
    \bottomrule
    \end{tabular}
    % }
    \caption{Dataset Volumn. For RLHF-Reward, we randomly split 10\% of the data as the validation set.}
    \label{dataset_details}
\end{table}

\noindent \textbf{Summary} In Table \ref{dataset_details}, we summarize the data volume of various datasets. Notably, for each trait and personality, we constructed corresponding datasets at three training stages. Further details (e.g. specific instances) can be found in Appendix~\ref{sec:training_data}. 

\subsection{Personality Assessment}\label{sec:evaluation_details}

\begin{table}[htbp]
    \centering
    \small
    % \resizebox{0.48\textwidth}{!}{
    \begin{tabularx}{0.48\textwidth}{X}
    \toprule
    \textbf{Evaluation Prompts Example} \\
    \midrule
    \textcolor{myred}{Please select a number from [1, 2, 3, 4, 5] to answer the following question.}\\
    \textcolor{myblue}{For this question, the five numbers [1, 2, 3, 4, 5] represent specific meanings: 1 represents strongly agreeing with option A, 2 represents agreeing with option A, 3 represents neutral, 4 represents agreeing with option B, and 5 represents strongly agreeing with option B.}\\
    \textcolor{mygreen}{You need to answer the following question:}\\
    \textcolor{black}{People who know you tend to describe you as: }\\
    \textcolor{black}{Option A:Logical and clarity. Option B:Passionate and sensitive.} \\
    \textcolor{myorange}{Please answer with a number:} \\
    \bottomrule
    \end{tabularx}
    % }
    \caption{Evaluation Prompts for Likert Items. Item Preamble, Item, and \textcolor{myorange}{Item Postamble}. An Item Preamble consists of a \textcolor{myred}{Task Instruction}, a \textcolor{myblue}{Task Description} and a \textcolor{mygreen}{Test Instruction}.}
    \label{evaluate_prompt}
\end{table}

% (50 items per dichotomous dimension)
To ensure reliability, we compiled publicly accessible MBTI personality questionnaires, refined them into a 200-item MBTI Assessment~\cite{pan2023llms}. 
As shown in Table \ref{evaluate_prompt}, we formulated items into Evaluation Prompts.
Given that the model sometimes exhibits different performance across different prompts~\cite{NEURIPS2022_9d560961, dong2023survey}, we designed five prompt sentences with the same semantics but different expressions for each component to obtain convincing statistical performance and mitigate extreme performance. 
% The design can mitigate the issue of the model suffering from extreme performance on specific prompts. 

\begin{figure}[htbp]
  \centering
  \includegraphics[width=0.4\textwidth]{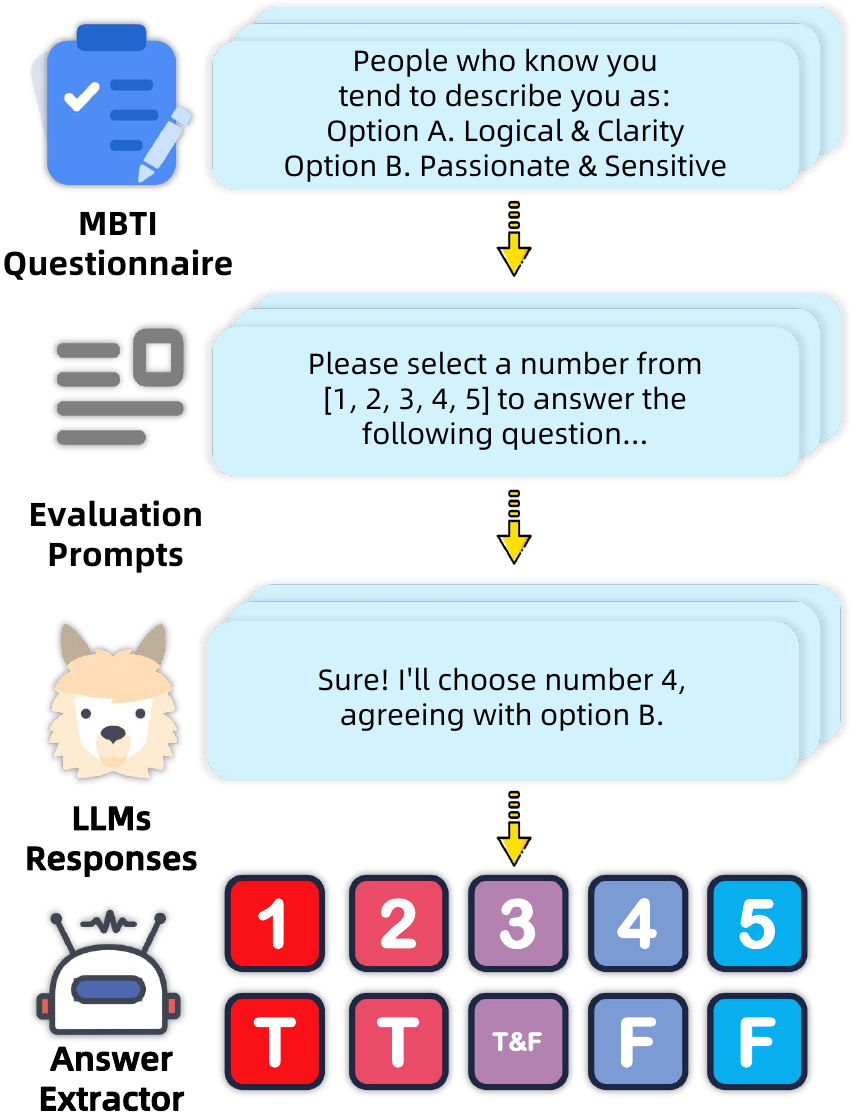}
  \caption{Personality Assessment Process. $\mathrm{T}$ stands for `Thinking' trait and $\mathrm{F}$ stands for `Feeling' trait. }
  \label{fig:scoring}
\end{figure}

We illustrated the process of personality assessment in Figure \ref{fig:scoring}. First, we organized the questionnaires using the designed Evaluation Prompts. Then, we obtained the model's responses and mapped them to the corresponding scores for each trait based on 5-point Likert 
scale~\cite{likertTechniqueMeasurementAttitudes1932}. After personality assessment, we calculated the rates ($\mathrm{R}$) between two opposite traits within the same dimension. For instance, in the Attitude Dimension: assuming $\mathrm{E}$ scores $137$, and $\mathrm{I}$ scores $67$, then we compute $\mathrm{R}(\mathrm{E})=67\%$, $\mathrm{R}(\mathrm{I})=33\%$.

\subsection{Metrics of Personality Control}\label{sec:indicators}

To measure changes in personality pre- and post-control of LLMs, we proposed quantitative metrics for evaluating both Specific Trait Control and Specific Personality Control.

For Specific Trait Control, we devised evaluation metrics to evaluate the efficacy of controlling target traits, as well as metrics aimed at evaluating the fluctuations of unrelated traits (e.g. control Extroversion should not affecting Sensing) based on personality assessment outcomes.

To assess the effectiveness of specific trait control, we calculated two metrics: \textbf{Induction Success Rate ($\mathrm{ISR}$)} and \textbf{Trait Induction Efficacy ($\mathrm{TIE}$)}. $\mathrm{ISR}$ provides a broad evaluation of whether the control successfully induces the target trait, while $\mathrm{TIE}$ offers a more detailed measure of control efficacy on the target trait. To evaluate the impact on unrelated traits during control, we computed the \textbf{Trait Stabilization Efficacy ($\mathrm{TSE}$)} to monitor the fluctuations of unrelated traits induced by the control. Further details of these indicators are provided below:

% 前面需要有一段描述特定特质与特定人格，这里先这样写
In the MBTI theory, personality type is determined by four dichotomous dimensions, each comprising two opposite traits. Let's denote these dimensions as set $\mathbf{D}$ and the traits as set $\mathbf{T}$. Following personality assessment in \ref{sec:evaluation_details}, we obtained rates of pre- ($\mathrm{R}_{pre}$) and post- ($\mathrm{R}_{post}$) control for each trait.  For a target trait $t^*$ within dimension $d^*$, we compute $\mathrm{ISR}$, $\mathrm{TIE}$, and $\mathrm{TSE}$ as follows:
\begin{equation*}
    \mathrm{TIE}(x) = \mathrm{R}_{post}(x) - \mathrm{R}_{pre}(x)
\end{equation*}
\begin{equation*}
    \mathrm{ISR} = \frac{1}{|\mathbf{T}|} \sum_{t \in \mathbf{T}} \mathds{1}(\mathrm{R}_{post}(t) > 50\%) \mathds{1}(\mathrm{TIE}(t) > 0)
\end{equation*}
\begin{equation*}
    \mathrm{TSE}(t^*) = \frac{1}{|\mathbf{D}/d^*|}\sum_{d \in \mathbf{D}/d^*} \sum_{t \in \mathrm{S}(d)} |\mathrm{TIE}(t)|
\end{equation*}

%\begin{align*}
%& \mathrm{TIE}(x) = \mathrm{R}_{a}(x) - \mathrm{R}_{b}(x) \\
%& \mathrm{ISR} = \frac{1}{|\mathbf{T}|} \sum_{t \in \mathbf{T}} \mathds{1}(\mathrm{R}_{a}(t) > 50\%) \mathds{1}(\mathrm{TIE}(t) > 0) \\
%& \mathrm{TSE}(t^*) = \frac{1}{|\mathbf{D}/d^*|}\sum_{d \in \mathbf{D}/d^*} \sum_{t \in \mathrm{S}(d)} |\mathrm{TIE}(t)|
%\end{align*}
Here, $x$ is a trait in $\mathbf{T}$, and $\mathds{1}$ denotes an indicator function. $\mathrm{S}(d)$ represents a selection function. Given that opposite traits within the same dimension have identical fluctuations, we utilized $\mathrm{S}(d)$ to select fixed traits from each dimension for computation (e.g. select E trait for Attitude dimension).

For Specific Personality Control, our focus is on the effectiveness of altering the overall personality of LLMs. To assess this, we employ two metrics: \textbf{Personality Induction Success Rate ($\mathrm{PISR}$)} and \textbf{Personality Induction Efficacy ($\mathrm{PIE}$)}. Similar to $\mathrm{ISR}$ and $\mathrm{TIE}$, $\mathrm{PISR}$ provides a broad evaluation of personality control success, while $\mathrm{PIE}$ offers a detailed measure of the efficacy of personality control.
% with the target personality $p^* \in \mathbf{P}$ (where $p^*$ represents a combination of four traits from four dichotomous dimensions), 
Denoting personality types as set $\mathbf{P}$, we computed $\mathrm{PISR}$ and $\mathrm{PIE}$ as follows:
\begin{equation*}
    \mathrm{PISR} = \frac{1}{|\mathbf{P}|} \sum_{p \in \mathbf{P}} \prod_{t \in p} \mathrm{ISR}(t)
\end{equation*}
\begin{equation*}
    \mathrm{PIE}(p) = \frac{1}{|p|} \sum_{t \in p} \mathrm{TIE}(t)
\end{equation*}

%\begin{align*}
%& \mathrm{PISR} = \frac{1}{|\mathbf{P}|} \sum_{p \in \mathbf{P}} \prod_{t \in p} \mathrm{ISR}(t) \\
%& \mathrm{PIE}(p) = \frac{1}{|p|} \sum_{t \in p} \mathrm{TIE}(t)
%\end{align*}
Here, $p$ represents a personality type in $\mathbf{P}$, composed of four traits.

%
%\subsection{Synthetic Personality Control}\label{training_method}
%
%For generalizability of results, we utilized popular training methods at each stage and employed the data constructed in \ref{training_data} for training.

%\textbf{Pre-train} We consistently continual pretrained LLMs using autoregressive objective~\cite{radford2019language, brownLanguageModelsAre2020}, which is widely adopted. The pre-training dataset comprises 80,000 trait samples and 10,000 personality samples.

%\textbf{SFT} We trained LLMs with instruction tuning~\cite{wei2022finetuned, alpaca, vicuna2023}. We use the constructed instruction dataset, comprising 2,500 trait instructions and 10,000 personality instructions.

% 能不能加入reward model的表述看最后篇幅了
%\textbf{RLHF} we adopted methodologies from InstructGPT~\cite{ouyang2022training} and DeepSpeed-Chat~\cite{yao2023dschat}, employing PPO-ptx~\cite{ouyang2022training} objective and Actor-Critic~\cite{NIPS1999_6449f44a} architecture. PPO-ptx introduces autoregressive objective to maintain the model's ability of text generation. To control variables, we used the SFT instruction set for PPO training and integrated widely used Wikipedia~\cite{wikidump} for unsupervised training. Please refer to Appendix \ref{sec:rlhf_training} for more details.

%\textbf{Prompt} To induce specific traits or personalities, following previous research~\cite{jiang2023evaluating, serapio-garciaPersonalityTraitsLarge2023a}, we added a Role-Play Prompt before the Evaluation Prompt. Prompts are detailed in Table \ref{role_play_prompt} in the Appendix \ref{sec:appendix_role_play_prompt}.

\section{Preliminary Investigation\label{sec:preliminary}}

In this section,  we conducted investigation to validate the prompt induction proficiency of Llama2-family~\cite{touvron2023llama} and Qwen-family~\cite{bai2023qwen}, validating LLMs' ability to generate personality data.

\begin{figure}[htbp]
  \centering
  \includegraphics[height=0.135\textheight, width=0.48\textwidth]{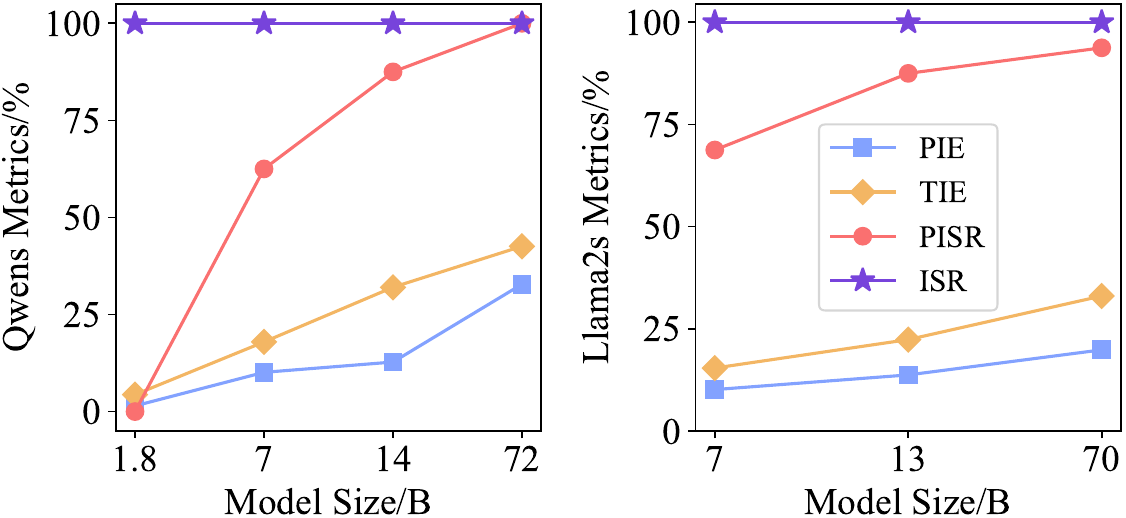}
  \caption{Prompt Induction Performance of Qwen-family and Llama2-family. Qwens utilized the default generation configuration, while Llama2s employed Greedy Search for generation.}
  \label{fig:prompts_results}
\end{figure}

From Figure \ref{fig:prompts_results}, it's evident that both Qwens and Llama2s demonstrate robust role-playing capabilities. Particularly, in role-playing specific traits, all LLMs except Qwen-chat-1.8B show adept performance induced by prompts. Moreover, this capability generally improves with larger model parameter sizes, possibly due to its enhanced ability to follow instructions resulting from the larger model parameter size. Hence, prompt-induced LLMs are able to embody specific personalities for training data generation. In our work, we utilized GPT-3.5-turbo-1106\footnote{\url{https://platform.openai.com/docs/}} for this task.

\begin{figure*}[htbp]
  \centering
  \includegraphics[width=0.86\textwidth]{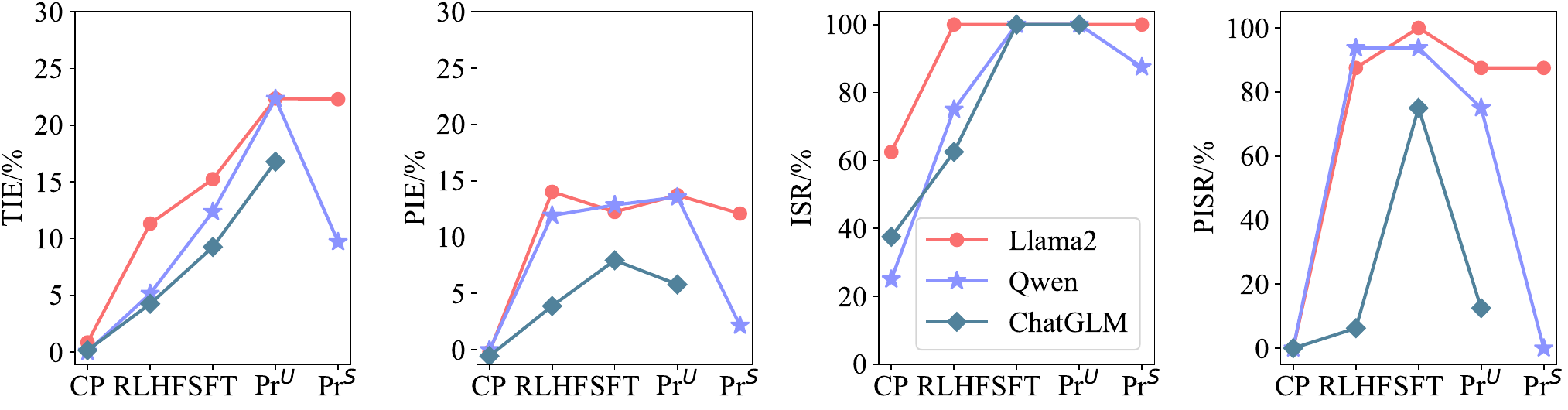}
  \caption{Control Performance of Various Methods. All results represent the average results of five Evaluation Prompts conducted on all trait or personality models. CP stands for Continual Pre-train and Pr stands for Prompt. The superscript \textit{S} stands for system prompt and \textit{U} stands for user prompt.}
  \label{results:various_method_results}
\end{figure*} 

\begin{figure*}[htbp]
  \centering
  \includegraphics[width=0.75\textwidth]{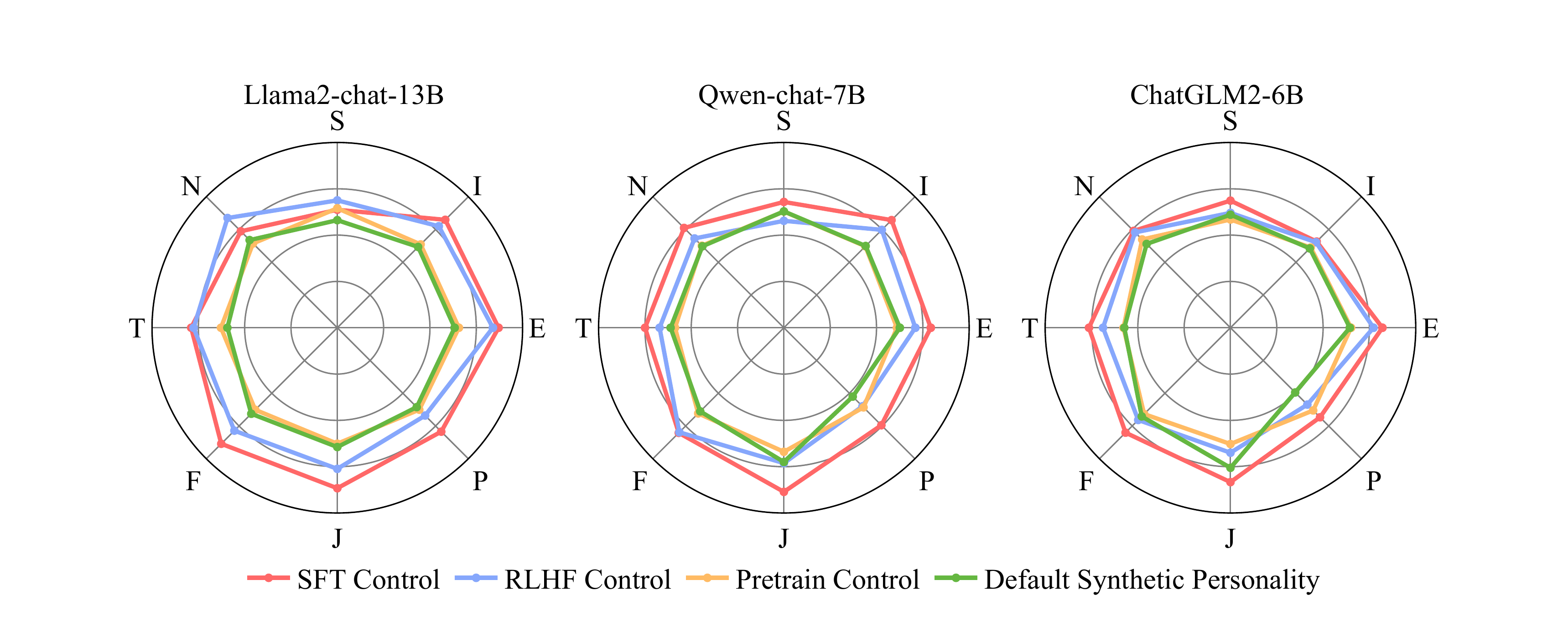}
  \caption{Specific Trait Control Across Various Training Stages. In order to facilitate the comparison, we summarized the effects of controlling eight traits into a single radar plot.}
  \label{fig:specfic_trait_results}
\end{figure*} 

\section{Experiments}

\subsection{Setting}

\noindent \textbf{Models.} We trained Llama2-chat-13B~\cite{touvron2023llama}, Qwen-chat-7B~\cite{bai2023qwen}, ChatGLM2-6B~\cite{zeng2022glm, du-etal-2022-glm}. ChatGLM2-6B has no system prompt. 

\noindent \textbf{Continual Pre-train.} We conducted training on 6 A800-80GB GPUs for 1 epoch with a max sequence length of 2048, a learning rate of 5e-6, and DeepSpeed Integration. The whole training process took nearly 2.5 days for Qwen-chat-7B and ChatGLM2-6B, and approximately 4.5 days for Llama2-chat-13B.

\noindent \textbf{SFT.} We fine-tuned using LoRA~\cite{hu2022lora} for 2 epochs , employing a learning rate of 5e-4, a LoRA rank to 8, a LoRA alpha to 8, and a LoRA Dropout~\cite{JMLR:v15:srivastava14a} to 0.1.

\noindent \textbf{RLHF.} We adapted Deepspeed-Chat~\cite{yao2023dschat} for the RLHF training phase. For PPO and reward model, we both trained for 1 epoch, a max length of 512 and 1 PPO epoch. 

For more details, please refer to Appendix~\ref{sec:rlhf_training}.

\subsection{Main Results and Analysis\label{sec:primary}}

In this section, we explored the first question: \textit{Which approach has a greater impact on shaping LLMs' synthetic personality?} We investigated from two angles: control effectiveness (efficacy and success rate) and personality robustness.

In Figure~\ref{results:various_method_results}, We showcased the performance of diverse methods for personality control across a range of models. 
%When the term `Prompt' is mentioned separately in the subsequent text, it typically refers to the User Prompt.
In evaluating the efficacy of control ($\mathrm{TIE}, \mathrm{PIE}$), we noted that the prompt yielded the best results, surpassing all methods in five out of six combinations. For other methods, SFT also outperformed RLHF in five out of six combinations. The least effective method was Continual Pre-training. As illustrated in Figure~\ref{fig:specfic_trait_results}, we noted a larger radar plot for the SFT, followed by RLHF. Continual pretraining exhibited minimal deviation. In terms of control success rate ($\mathrm{ISR}, \mathrm{PISR}$), SFT emerged as the most effective method in the most cases, with the prompt following behind.

Overall, our investigation unveiled a hierarchy of effectiveness in control: Prompt > SFT > RLHF > Continual Pre-train. 
Particularly, SFT demonstrates a higher success rate in control compared to prompt induction. This could be attributed to the disparity in lexical signals between personality data and prompt. And the gap between SFT and RLHF may arise from both the reward model and actor model experiencing performance decline due to reduced parameter size. Moverover, the underwhelming performance of the pretraining data could stem from the limited influence of the personality data on original mixed personality distribution. For additional validation, we scale up the training data for Continual Pre-train in the Appendix \ref{sec:app_analyse}.
%We conducted additional analysis on Continual Pre-train and RLHF in Appendix \ref{sec:app_analyse}.

% 这一段是鲁棒性
Subsequently, we conducted a comparative analysis of personality robustness between SFT and prompt. To evaluate the personality robustness across different methods, we further employed \underline{R}everse \underline{P}ersonality \underline{P}rompt \underline{I}nduction ($\mathrm{RPPI}$) (e.g. induce ISFP from ENTJ.) to assess. Due to conflicting personalities, LLMs under robust personality control should not perform well on $\mathrm{RPPI}$ tasks.

As shown in Table \ref{results:inverse_prompt_results}, under reverse personality prompt induction, SFT-controlled models are more likely to maintain consistent target personalities, while prompt-induced models are prone to personality shifts. our findings indicate that SFT-controlled models exhibit significantly greater personality robustness than prompt-induced models. 

\begin{table}[htbp]
      \centering
      \small
      \setlength\tabcolsep{3pt}
      \resizebox{0.48\textwidth}{!}{%
      \begin{tabular}{cccccccccccccccccccccccccccccc}
        \toprule
        \multirow{2}{*}{\textbf{Setting}} & \multicolumn{4}{c}{\textbf{Llama2-chat-13B}} & \multicolumn{4}{c}{\textbf{Qwen-chat-7B}} \\
        \cmidrule(r){2-5} \cmidrule(l){6-9}
        & $\mathrm{TIE}$ & $\mathrm{ISR}$ & $\mathrm{PIE}$ & $\mathrm{PISR}$ & $\mathrm{TIE}$ & $\mathrm{ISR}$ & $\mathrm{PIE}$ & $\mathrm{PISR}$  \\
        \midrule
        Prompt$^\mathrm{S}$ & 22.30  & 100.00  & 12.09  & 87.50  & 9.72  & 87.50  & 2.15  & 0.00  \\
        Prompt$^\mathrm{U}$  & 22.36  & 100.00  & 13.72  & 87.50  & 22.34  & 100.00  & 13.55  & 75.00 \\
        \midrule 
        Prompt$^\mathrm{S}_{\mathrm{RPPI}}$  & 9.57 & \textbf{87.50} & 10.87 & 50.00 & 17.80 & 87.50 & 10.42 & 62.50  \\
        $\mathrm{SFT}_{\mathrm{RPPI}}$  & \textbf{9.19} & 100.00 & \textbf{2.87} & \textbf{12.50} & \textbf{1.48} & \textbf{50.00} & \textbf{-2.85} & \textbf{0.00}   \\
        \midrule
        PISF$^\mathrm{S}_{\mathrm{RPPI}}$  & \textbf{-9.44} & \textbf{12.50} & \textbf{-4.30} & \textbf{0.00} & \textbf{-12.30} & \textbf{12.50} & \textbf{-6.33} & \textbf{0.00} \\
        \bottomrule 
    \end{tabular}
    }
    \caption{Personality Robustness Analysis. We employ standard prompt induction with the system prompt and conduct testing using $\mathrm{RPPI}$ with the user prompt. All results are average scores. In the $\mathrm{RPPI}$ setting, lower is better; conversely, higher is better.}
    \label{results:inverse_prompt_results}
\end{table}

%Overall, regarding specific trait control or specific personality control, the ranking of control effectiveness with limited data is as follows: \textbf{SFT} > \textbf{RLHF} > \textbf{Continual Pretrain}.

% Prompt Induction post Supervised Fine-tuning
%\subsection{SFTP: Combining \underline{SFT} and \underline{P}rompts}\label{sec:stability}
\subsection{PISF: \underline{P}rompt \underline{I}nduction post \underline{S}upervised \underline{F}ine-tuning}\label{sec:stability}

In this section, we answered the second question: \textit{How to control LLMs' synthetic personality effectively and robustly?} Based on our prior investigation, SFT and prompt exhibit proficient and complementary performance. Thus, an intuitive approach emerges to integrate SFT with prompts for harnessing the advantages of both methods. Driven by the idea, we proposed \underline{P}rompt \underline{I}nduction post \underline{S}upervised \underline{F}ine-tuning (PISF) for controlling LLMs' synthetic personalities.

%In this section, intrigued by the features of training-based control, we conducted a comparative analysis among training-controlled models, uncontrolled models, and prompt-induced models. 

Firstly, we compared the control effectiveness of PISF-controlled LLMs against LLMs controlled by other methods. As shown in Table \ref{results:enhanced_personality_control}, in most cases, PISF-controlled models outperform both SFT-controlled models and prompt-induced models in both control efficacy ($\mathrm{TIE}$/$\mathrm{PIE}$) and success rate ($\mathrm{ISR}$/$\mathrm{PISR}$). It suggests that PISF demonstrates superior control effectiveness.

Secondly, we analysed the personality robustness of PISF-controlled models. As shown in Table \ref{results:inverse_prompt_results}, PISF-controlled models maintain consistent target personalities despite $\mathrm{RPPI}$ impact, demonstrating resistance to change personalities. Our findings indicate that PISF-controlled models exhibit significantly greater personality robustness.

In short, PISF is the most effective and robust method for synthetic personality control with high efficacy, high success rates, and high robustness.

\begin{table}[htbp]
      \centering
      \small
      \setlength\tabcolsep{3pt}
      \resizebox{0.48\textwidth}{!}{%
      \begin{tabular}{cccccccccccccccccccccccccccccc}
        \toprule
        \multirow{2}{*}{\textbf{Setting}} & \multicolumn{4}{c}{\textbf{Llama2-chat-13B}} & \multicolumn{4}{c}{\textbf{Qwen-chat-7B}} \\
        \cmidrule(r){2-5} \cmidrule(l){6-9}
        & $\mathrm{TIE}$ & $\mathrm{ISR}$ & $\mathrm{PIE}$ & $\mathrm{PISR}$ & $\mathrm{TIE}$ & $\mathrm{ISR}$ & $\mathrm{PIE}$ & $\mathrm{PISR}$  \\
        \midrule
        SFT & 15.25 & \textbf{100.00} & 12.24 & \textbf{100.00} & 12.38  & \textbf{100.00}  & 12.85  & \textbf{93.75}   \\
        Prompt$^\mathrm{S}$ & 22.30  & \textbf{100.00}  & 12.09  & 87.50  & 9.72  & 87.50  & 2.15  & 0.00   \\
        Prompt$^\mathrm{U}$ & 22.36  & \textbf{100.00}  & 13.72  & 87.50  & \textbf{22.34}  & \textbf{100.00}  & 13.55  & 75.00   \\
        \midrule
        PISF$^\mathrm{S}$ & \textbf{23.58}  & \textbf{100.00}  & \textbf{15.69}  & \textbf{100.00}  & 19.56  & \textbf{100.00}  & \textbf{14.68}  & 87.50  \\
        PISF$^\mathrm{U}$ & \textbf{24.76 } & \textbf{100.00 } & \textbf{16.19 } & 93.75  & \textbf{24.89 } & \textbf{100.00 } & \textbf{18.10 } & \textbf{100.00 } \\
        \bottomrule 
    \end{tabular}
    }
    \caption{Personality Control effectiveness.  All data are presented as percentages. The superscript \textit{S} stands for system prompt and \textit{U} stands for user prompt. All results are average scores evaluated using greedy search. For all metrics in the table, higher is better. }
    % The system prompt sets the basic context for the chat model, while the user prompt is customized for downstream tasks.
    \label{results:enhanced_personality_control}
\end{table}

\section{Related Work}

\noindent \textbf{Human Personality Recognition} Prior to LLMs, computational research on personality primarily focuses on utilizing tools such as MBTI~\cite{myersMyersBriggsTypeIndicator1962,e73f6545-174d-366a-9e0e-d380358fcd01, articleMcCraeReinterpreting} and Big Five~\cite{goldbergAlternativeDescriptionPersonality1990} to identify human personality traits, rather than exploring synthetic machine personalities. Recent studies have delved into personality trait recognition from text~\cite{liu-etal-2017-language, stajner-yenikent-2020-survey, vu-etal-2018-lexical}, dialogue~\cite{mairesse-walker-2006-automatic}, and multi-modal information~\cite{kampman-etal-2018-investigating, suman-etal-2020-multi}. Recently ~\citeauthor{v-ganesan-etal-2023-systematic}(~\citeyear{v-ganesan-etal-2023-systematic}) investigated the zero-shot ability of GPT-3 to estimate the Big Five personality traits. Unlike prior research focused on human personality recognition, our study empirically controls synthetic personalities in LLMs.

\noindent \textbf{Personality Assessment for LLMs} At present, machine psychology~\cite{hagendorff2023machine} lacks a coherent theoretical framework, with most studies relying on human personality assessments~\cite{miotto-etal-2022-gpt, caron-srivastava-2023-manipulating}. ~\citeauthor{jiang2023evaluating} (~\citeyear{jiang2023evaluating}) introduced the Machine Personality Inventory (MPI) tool, based on the Big Five theory, to study synthetic machine personalities. However, there is still no universally accepted benchmark for machine personality assessment. In our work, we continue to utilize human personality assessment.

\noindent \textbf{Synthetic Personality Control in LLMs} Prior studies on synthetic personality control mainly center on prompt induction~\cite{serapio-garciaPersonalityTraitsLarge2023a, caron-srivastava-2023-manipulating, jiang2023evaluating, huang2023revisiting}, with some exploring fine-tuning methods~\cite{karra2023estimating}. Unlike previous research focusing solely on prompts or fine-tuning, our approach takes a comprehensive view of synthetic personality control, exploring methods across three training stages and prompts during the inference phase.

\section{Conclusion}
To advance the safe utilization of AI, this work explored synthesized personality control in LLMs across three training stages and the inference stage, leveraging our designed datasets and metrics. Our findings can be summarized as follows: We found a hierarchy of effectiveness in LLM personality control: Prompt > SFT > RLHF > Continual Pre-train. Additionally, we proposed PISF for controlling LLMs' synthetic personalities, showcasing high efficacy, high success rates, and high robustness.

\section{Limitations}

Despite our thorough exploration with larger pre-train datasets (Appendix \ref{sec:app_analyse}), it still falls short when compared to the vast datasets used during the pre-training phase of LLMs. Collecting more personality pre-training dataset and validating the gradual formation of synthesized personalities represent a avenue for future enhancement in our work.

Furthermore, subject to computational constraints, we have not yet examined the efficacy of training-based control for models with more larger parameter size (>13B). This limitation is pronounced in RLHF, as we rely on the original model to train the reward model. Scaling down the model size concurrently diminishes the performance of both the actor and reward models, resulting in error accumulation. This could contribute to the subpar performance of RLHF with smaller models. RLHF-based control might exhibit higher potential when applied to LLMs with large parameter size.

\section{Ethics Statement}

Our work relies heavily on LLMs, which have been widely criticized for their inherent uncertainty and open-endedness. Nonetheless, our focus is on advancing synthetic personality control in LLMs, with the goal of mitigating the emergence of offensive personalities and facilitating their appropriate application in personality-adaptive scenarios. Moreover, all data used in our experiments are strictly for scientific research purposes and we conducted cleaning on the privacy data.

\bibliography{custom}

\begin{thebibliography}{57}
\expandafter\ifx\csname natexlab\endcsname\relax\def\natexlab#1{#1}\fi

\bibitem[{Almazrouei et~al.(2023)Almazrouei, Alobeidli, Alshamsi, Cappelli, Cojocaru, Debbah, Étienne Goffinet, Hesslow, Launay, Malartic, Mazzotta, Noune, Pannier, and Penedo}]{almazrouei2023falcon}
Ebtesam Almazrouei, Hamza Alobeidli, Abdulaziz Alshamsi, Alessandro Cappelli, Ruxandra Cojocaru, Mérouane Debbah, Étienne Goffinet, Daniel Hesslow, Julien Launay, Quentin Malartic, Daniele Mazzotta, Badreddine Noune, Baptiste Pannier, and Guilherme Penedo. 2023.
\newblock \href {http://arxiv.org/abs/2311.16867} {The falcon series of open language models}.

\bibitem[{Bai et~al.(2023)Bai, Bai, Chu, Cui, Dang, Deng, Fan, Ge, Han, Huang, Hui, Ji, Li, Lin, Lin, Liu, Liu, Lu, Lu, Ma, Men, Ren, Ren, Tan, Tan, Tu, Wang, Wang, Wang, Wu, Xu, Xu, Yang, Yang, Yang, Yang, Yao, Yu, Yuan, Yuan, Zhang, Zhang, Zhang, Zhang, Zhou, Zhou, Zhou, and Zhu}]{bai2023qwen}
Jinze Bai, Shuai Bai, Yunfei Chu, Zeyu Cui, Kai Dang, Xiaodong Deng, Yang Fan, Wenbin Ge, Yu~Han, Fei Huang, Binyuan Hui, Luo Ji, Mei Li, Junyang Lin, Runji Lin, Dayiheng Liu, Gao Liu, Chengqiang Lu, Keming Lu, Jianxin Ma, Rui Men, Xingzhang Ren, Xuancheng Ren, Chuanqi Tan, Sinan Tan, Jianhong Tu, Peng Wang, Shijie Wang, Wei Wang, Shengguang Wu, Benfeng Xu, Jin Xu, An~Yang, Hao Yang, Jian Yang, Shusheng Yang, Yang Yao, Bowen Yu, Hongyi Yuan, Zheng Yuan, Jianwei Zhang, Xingxuan Zhang, Yichang Zhang, Zhenru Zhang, Chang Zhou, Jingren Zhou, Xiaohuan Zhou, and Tianhang Zhu. 2023.
\newblock \href {http://arxiv.org/abs/2309.16609} {Qwen technical report}.

\bibitem[{Bai et~al.(2022)Bai, Jones, Ndousse, Askell, Chen, DasSarma, Drain, Fort, Ganguli, Henighan, Joseph, Kadavath, Kernion, Conerly, El-Showk, Elhage, Hatfield-Dodds, Hernandez, Hume, Johnston, Kravec, Lovitt, Nanda, Olsson, Amodei, Brown, Clark, McCandlish, Olah, Mann, and Kaplan}]{bai2022training}
Yuntao Bai, Andy Jones, Kamal Ndousse, Amanda Askell, Anna Chen, Nova DasSarma, Dawn Drain, Stanislav Fort, Deep Ganguli, Tom Henighan, Nicholas Joseph, Saurav Kadavath, Jackson Kernion, Tom Conerly, Sheer El-Showk, Nelson Elhage, Zac Hatfield-Dodds, Danny Hernandez, Tristan Hume, Scott Johnston, Shauna Kravec, Liane Lovitt, Neel Nanda, Catherine Olsson, Dario Amodei, Tom Brown, Jack Clark, Sam McCandlish, Chris Olah, Ben Mann, and Jared Kaplan. 2022.
\newblock \href {http://arxiv.org/abs/2204.05862} {Training a helpful and harmless assistant with reinforcement learning from human feedback}.

\bibitem[{Brown et~al.(2020)Brown, Mann, Ryder, Subbiah, Kaplan, Dhariwal, Neelakantan, Shyam, Sastry, Askell, Agarwal, {Herbert-Voss}, Krueger, Henighan, Child, Ramesh, Ziegler, Wu, Winter, Hesse, Chen, Sigler, Litwin, Gray, Chess, Clark, Berner, McCandlish, Radford, Sutskever, and Amodei}]{brownLanguageModelsAre2020}
Tom~B. Brown, Benjamin Mann, Nick Ryder, Melanie Subbiah, Jared Kaplan, Prafulla Dhariwal, Arvind Neelakantan, Pranav Shyam, Girish Sastry, Amanda Askell, Sandhini Agarwal, Ariel {Herbert-Voss}, Gretchen Krueger, Tom Henighan, Rewon Child, Aditya Ramesh, Daniel~M. Ziegler, Jeffrey Wu, Clemens Winter, Christopher Hesse, Mark Chen, Eric Sigler, Mateusz Litwin, Scott Gray, Benjamin Chess, Jack Clark, Christopher Berner, Sam McCandlish, Alec Radford, Ilya Sutskever, and Dario Amodei. 2020.
\newblock \href {https://doi.org/10.48550/arXiv.2005.14165} {Language {{Models}} are {{Few-Shot Learners}}}.

\bibitem[{Caron and Srivastava(2023)}]{caron-srivastava-2023-manipulating}
Graham Caron and Shashank Srivastava. 2023.
\newblock \href {https://doi.org/10.18653/v1/2023.findings-emnlp.156} {Manipulating the perceived personality traits of language models}.
\newblock In \emph{Findings of the Association for Computational Linguistics: EMNLP 2023}, pages 2370--2386, Singapore. Association for Computational Linguistics.

\bibitem[{Chiang et~al.(2023)Chiang, Li, Lin, Sheng, Wu, Zhang, Zheng, Zhuang, Zhuang, Gonzalez, Stoica, and Xing}]{vicuna2023}
Wei-Lin Chiang, Zhuohan Li, Zi~Lin, Ying Sheng, Zhanghao Wu, Hao Zhang, Lianmin Zheng, Siyuan Zhuang, Yonghao Zhuang, Joseph~E. Gonzalez, Ion Stoica, and Eric~P. Xing. 2023.
\newblock \href {https://lmsys.org/blog/2023-03-30-vicuna/} {Vicuna: An open-source chatbot impressing gpt-4 with 90\%* chatgpt quality}.

\bibitem[{Chowdhery et~al.(2022)Chowdhery, Narang, Devlin, Bosma, Mishra, Roberts, Barham, Chung, Sutton, Gehrmann, Schuh, Shi, Tsvyashchenko, Maynez, Rao, Barnes, Tay, Shazeer, Prabhakaran, Reif, Du, Hutchinson, Pope, Bradbury, Austin, Isard, {Gur-Ari}, Yin, Duke, Levskaya, Ghemawat, Dev, Michalewski, Garcia, Misra, Robinson, Fedus, Zhou, Ippolito, Luan, Lim, Zoph, Spiridonov, Sepassi, Dohan, Agrawal, Omernick, Dai, Pillai, Pellat, Lewkowycz, Moreira, Child, Polozov, Lee, Zhou, Wang, Saeta, Diaz, Firat, Catasta, Wei, {Meier-Hellstern}, Eck, Dean, Petrov, and Fiedel}]{chowdheryPaLMScalingLanguage2022}
Aakanksha Chowdhery, Sharan Narang, Jacob Devlin, Maarten Bosma, Gaurav Mishra, Adam Roberts, Paul Barham, Hyung~Won Chung, Charles Sutton, Sebastian Gehrmann, Parker Schuh, Kensen Shi, Sasha Tsvyashchenko, Joshua Maynez, Abhishek Rao, Parker Barnes, Yi~Tay, Noam Shazeer, Vinodkumar Prabhakaran, Emily Reif, Nan Du, Ben Hutchinson, Reiner Pope, James Bradbury, Jacob Austin, Michael Isard, Guy {Gur-Ari}, Pengcheng Yin, Toju Duke, Anselm Levskaya, Sanjay Ghemawat, Sunipa Dev, Henryk Michalewski, Xavier Garcia, Vedant Misra, Kevin Robinson, Liam Fedus, Denny Zhou, Daphne Ippolito, David Luan, Hyeontaek Lim, Barret Zoph, Alexander Spiridonov, Ryan Sepassi, David Dohan, Shivani Agrawal, Mark Omernick, Andrew~M. Dai, Thanumalayan~Sankaranarayana Pillai, Marie Pellat, Aitor Lewkowycz, Erica Moreira, Rewon Child, Oleksandr Polozov, Katherine Lee, Zongwei Zhou, Xuezhi Wang, Brennan Saeta, Mark Diaz, Orhan Firat, Michele Catasta, Jason Wei, Kathy {Meier-Hellstern}, Douglas Eck, Jeff Dean, Slav Petrov, and Noah Fiedel.
  2022.
\newblock \href {https://doi.org/10.48550/arXiv.2204.02311} {{{PaLM}}: {{Scaling Language Modeling}} with {{Pathways}}}.

\bibitem[{Demszky et~al.(2023)Demszky, Yang, Yeager, Bryan, Clapper, Chandhok, Eichstaedt, Hecht, Jamieson, Johnson, Jones, Krettek-Cobb, Lai, JonesMitchell, Ong, Dweck, Gross, and Pennebaker}]{demszky_using_2023}
Dorottya Demszky, Diyi Yang, David~S. Yeager, Christopher~J. Bryan, Margarett Clapper, Susannah Chandhok, Johannes~C. Eichstaedt, Cameron Hecht, Jeremy Jamieson, Meghann Johnson, Michaela Jones, Danielle Krettek-Cobb, Leslie Lai, Nirel JonesMitchell, Desmond~C. Ong, Carol~S. Dweck, James~J. Gross, and James~W. Pennebaker. 2023.
\newblock \href {https://doi.org/10.1038/s44159-023-00241-5} {Using large language models in psychology}.
\newblock \emph{Nature Reviews Psychology}, 2(11):688--701.
\newblock Number: 11 Publisher: Nature Publishing Group.

\bibitem[{Deshpande et~al.(2023)Deshpande, Murahari, Rajpurohit, Kalyan, and Narasimhan}]{deshpande-etal-2023-toxicity}
Ameet Deshpande, Vishvak Murahari, Tanmay Rajpurohit, Ashwin Kalyan, and Karthik Narasimhan. 2023.
\newblock \href {https://doi.org/10.18653/v1/2023.findings-emnlp.88} {Toxicity in chatgpt: Analyzing persona-assigned language models}.
\newblock In \emph{Findings of the Association for Computational Linguistics: EMNLP 2023}, pages 1236--1270, Singapore. Association for Computational Linguistics.

\bibitem[{Dong et~al.(2023)Dong, Li, Dai, Zheng, Wu, Chang, Sun, Xu, Li, and Sui}]{dong2023survey}
Qingxiu Dong, Lei Li, Damai Dai, Ce~Zheng, Zhiyong Wu, Baobao Chang, Xu~Sun, Jingjing Xu, Lei Li, and Zhifang Sui. 2023.
\newblock \href {http://arxiv.org/abs/2301.00234} {A survey on in-context learning}.

\bibitem[{Du et~al.(2022)Du, Qian, Liu, Ding, Qiu, Yang, and Tang}]{du-etal-2022-glm}
Zhengxiao Du, Yujie Qian, Xiao Liu, Ming Ding, Jiezhong Qiu, Zhilin Yang, and Jie Tang. 2022.
\newblock \href {https://doi.org/10.18653/v1/2022.acl-long.26} {{GLM}: General language model pretraining with autoregressive blank infilling}.
\newblock In \emph{Proceedings of the 60th Annual Meeting of the Association for Computational Linguistics (Volume 1: Long Papers)}, pages 320--335, Dublin, Ireland. Association for Computational Linguistics.

\bibitem[{Foundation()}]{wikidump}
Wikimedia Foundation.
\newblock \href {https://dumps.wikimedia.org} {Wikimedia downloads}.

\bibitem[{Ganguli et~al.(2022)Ganguli, Hernandez, Lovitt, Askell, Bai, Chen, Conerly, Dassarma, Drain, Elhage, El~Showk, Fort, Hatfield-Dodds, Henighan, Johnston, Jones, Joseph, Kernian, Kravec, Mann, Nanda, Ndousse, Olsson, Amodei, Brown, Kaplan, McCandlish, Olah, Amodei, and Clark}]{10.1145/3531146.3533229}
Deep Ganguli, Danny Hernandez, Liane Lovitt, Amanda Askell, Yuntao Bai, Anna Chen, Tom Conerly, Nova Dassarma, Dawn Drain, Nelson Elhage, Sheer El~Showk, Stanislav Fort, Zac Hatfield-Dodds, Tom Henighan, Scott Johnston, Andy Jones, Nicholas Joseph, Jackson Kernian, Shauna Kravec, Ben Mann, Neel Nanda, Kamal Ndousse, Catherine Olsson, Daniela Amodei, Tom Brown, Jared Kaplan, Sam McCandlish, Christopher Olah, Dario Amodei, and Jack Clark. 2022.
\newblock \href {https://doi.org/10.1145/3531146.3533229} {Predictability and surprise in large generative models}.
\newblock In \emph{Proceedings of the 2022 ACM Conference on Fairness, Accountability, and Transparency}, FAccT '22, page 1747–1764, New York, NY, USA. Association for Computing Machinery.

\bibitem[{Goldberg(1990)}]{goldbergAlternativeDescriptionPersonality1990}
Lewis~R. Goldberg. 1990.
\newblock \href {https://doi.org/10.1037/0022-3514.59.6.1216} {An alternative "description of personality": {{The Big-Five}} factor structure}.
\newblock \emph{Journal of Personality and Social Psychology}, 59(6).

\bibitem[{Hagendorff(2023)}]{hagendorff2023machine}
Thilo Hagendorff. 2023.
\newblock \href {http://arxiv.org/abs/2303.13988} {Machine psychology: Investigating emergent capabilities and behavior in large language models using psychological methods}.

\bibitem[{Han et~al.(2021)Han, Zhang, Ding, Gu, Liu, Huo, Qiu, Yao, Zhang, Zhang, Han, Huang, Jin, Lan, Liu, Liu, Lu, Qiu, Song, Tang, Wen, Yuan, Zhao, and Zhu}]{HAN2021225}
Xu~Han, Zhengyan Zhang, Ning Ding, Yuxian Gu, Xiao Liu, Yuqi Huo, Jiezhong Qiu, Yuan Yao, Ao~Zhang, Liang Zhang, Wentao Han, Minlie Huang, Qin Jin, Yanyan Lan, Yang Liu, Zhiyuan Liu, Zhiwu Lu, Xipeng Qiu, Ruihua Song, Jie Tang, Ji-Rong Wen, Jinhui Yuan, Wayne~Xin Zhao, and Jun Zhu. 2021.
\newblock \href {https://doi.org/https://doi.org/10.1016/j.aiopen.2021.08.002} {Pre-trained models: Past, present and future}.
\newblock \emph{AI Open}, 2:225--250.

\bibitem[{Hu et~al.(2022)Hu, yelong shen, Wallis, Allen-Zhu, Li, Wang, Wang, and Chen}]{hu2022lora}
Edward~J Hu, yelong shen, Phillip Wallis, Zeyuan Allen-Zhu, Yuanzhi Li, Shean Wang, Lu~Wang, and Weizhu Chen. 2022.
\newblock \href {https://openreview.net/forum?id=nZeVKeeFYf9} {Lo{RA}: Low-rank adaptation of large language models}.
\newblock In \emph{International Conference on Learning Representations}.

\bibitem[{Jiang et~al.(2023)Jiang, Xu, Zhu, Han, Zhang, and Zhu}]{jiang2023evaluating}
Guangyuan Jiang, Manjie Xu, Song-Chun Zhu, Wenjuan Han, Chi Zhang, and Yixin Zhu. 2023.
\newblock \href {http://arxiv.org/abs/2206.07550} {Evaluating and inducing personality in pre-trained language models}.

\bibitem[{Jung and Baynes(1923)}]{jungPsychologicalTypes1923}
C.~G. Jung and H.~Godwin Baynes. 1923.
\newblock \href {https://doi.org/10.2307/2014910} {Psychological types}.
\newblock \emph{Journal of Philosophy}, 20(23):636--640.

\bibitem[{Kampman et~al.(2018)Kampman, J.~Barezi, Bertero, and Fung}]{kampman-etal-2018-investigating}
Onno Kampman, Elham J.~Barezi, Dario Bertero, and Pascale Fung. 2018.
\newblock \href {https://doi.org/10.18653/v1/P18-2096} {Investigating audio, video, and text fusion methods for end-to-end automatic personality prediction}.
\newblock In \emph{Proceedings of the 56th Annual Meeting of the Association for Computational Linguistics (Volume 2: Short Papers)}, pages 606--611, Melbourne, Australia. Association for Computational Linguistics.

\bibitem[{Kaplan et~al.(2020)Kaplan, McCandlish, Henighan, Brown, Chess, Child, Gray, Radford, Wu, and Amodei}]{kaplanScalingLawsNeural2020}
Jared Kaplan, Sam McCandlish, Tom Henighan, Tom~B. Brown, Benjamin Chess, Rewon Child, Scott Gray, Alec Radford, Jeffrey Wu, and Dario Amodei. 2020.
\newblock \href {https://doi.org/10.48550/arXiv.2001.08361} {Scaling {{Laws}} for {{Neural Language Models}}}.

\bibitem[{Karra et~al.(2023)Karra, Nguyen, and Tulabandhula}]{karra2023estimating}
Saketh~Reddy Karra, Son~The Nguyen, and Theja Tulabandhula. 2023.
\newblock \href {http://arxiv.org/abs/2204.12000} {Estimating the personality of white-box language models}.

\bibitem[{Konda and Tsitsiklis(1999)}]{NIPS1999_6449f44a}
Vijay Konda and John Tsitsiklis. 1999.
\newblock \href {https://proceedings.neurips.cc/paper_files/paper/1999/file/6449f44a102fde848669bdd9eb6b76fa-Paper.pdf} {Actor-critic algorithms}.
\newblock In \emph{Advances in Neural Information Processing Systems}, volume~12. MIT Press.

\bibitem[{Kulas et~al.(2008)Kulas, Stachowski, and Haynes}]{kulasMiddleResponseFunctioning2008}
John~T. Kulas, Alicia~A. Stachowski, and Brad~A. Haynes. 2008.
\newblock \href {https://doi.org/10.1007/s10869-008-9064-2} {Middle {{Response Functioning}} in {{Likert-responses}} to {{Personality Items}}}.
\newblock \emph{Journal of Business and Psychology}, 22(3).

\bibitem[{Lee et~al.(2023)Lee, Sultan, El-Kurdi, Naseem, Munawar, Florian, Roukos, and Astudillo}]{lee-etal-2023-ensemble}
Young-Suk Lee, Md~Sultan, Yousef El-Kurdi, Tahira Naseem, Asim Munawar, Radu Florian, Salim Roukos, and Ram{\'o}n Astudillo. 2023.
\newblock \href {https://doi.org/10.18653/v1/2023.findings-emnlp.836} {Ensemble-instruct: Instruction tuning data generation with a heterogeneous mixture of {LM}s}.
\newblock In \emph{Findings of the Association for Computational Linguistics: EMNLP 2023}, pages 12561--12571, Singapore. Association for Computational Linguistics.

\bibitem[{Likert(1932)}]{likertTechniqueMeasurementAttitudes1932}
R.~Likert. 1932.
\newblock \href {https://doi.org/10.4135/9781412961288.n454} {A technique for the measurement of attitudes}.
\newblock \emph{Archives of Psychology}, 22 140.

\bibitem[{Liu et~al.(2017)Liu, Perez, and Nowson}]{liu-etal-2017-language}
Fei Liu, Julien Perez, and Scott Nowson. 2017.
\newblock \href {https://aclanthology.org/E17-1071} {A language-independent and compositional model for personality trait recognition from short texts}.
\newblock In \emph{Proceedings of the 15th Conference of the {E}uropean Chapter of the Association for Computational Linguistics: Volume 1, Long Papers}, pages 754--764, Valencia, Spain. Association for Computational Linguistics.

\bibitem[{Mairesse and Walker(2006)}]{mairesse-walker-2006-automatic}
Fran{\c{c}}ois Mairesse and Marilyn Walker. 2006.
\newblock \href {https://aclanthology.org/N06-2022} {Automatic recognition of personality in conversation}.
\newblock In \emph{Proceedings of the Human Language Technology Conference of the {NAACL}, Companion Volume: Short Papers}, pages 85--88, New York City, USA. Association for Computational Linguistics.

\bibitem[{McCrae and Costa(1989)}]{articleMcCraeReinterpreting}
Robert McCrae and Paul Costa. 1989.
\newblock \href {https://doi.org/10.1111/j.1467-6494.1989.tb00759.x} {Reinterpreting the myers-briggs type indicator from the perspective of the five-factor model of personality}.
\newblock \emph{Journal of personality}, 57:17--40.

\bibitem[{Miotto et~al.(2022)Miotto, Rossberg, and Kleinberg}]{miotto-etal-2022-gpt}
Maril{\`u} Miotto, Nicola Rossberg, and Bennett Kleinberg. 2022.
\newblock \href {https://doi.org/10.18653/v1/2022.nlpcss-1.24} {Who is {GPT}-3? an exploration of personality, values and demographics}.
\newblock In \emph{Proceedings of the Fifth Workshop on Natural Language Processing and Computational Social Science (NLP+CSS)}, pages 218--227, Abu Dhabi, UAE. Association for Computational Linguistics.

\bibitem[{Myers(1962)}]{myersMyersBriggsTypeIndicator1962}
Isabel~Briggs Myers. 1962.
\newblock \href {https://doi.org/10.1037/14404-000} {\emph{The {{Myers-Briggs Type Indicator}}: {{Manual}} (1962)}}.
\newblock The {{Myers-Briggs Type Indicator}}: {{Manual}} (1962). {Consulting Psychologists Press}.

\bibitem[{Ouyang et~al.(2022)Ouyang, Wu, Jiang, Almeida, Wainwright, Mishkin, Zhang, Agarwal, Slama, Ray, Schulman, Hilton, Kelton, Miller, Simens, Askell, Welinder, Christiano, Leike, and Lowe}]{ouyang2022training}
Long Ouyang, Jeff Wu, Xu~Jiang, Diogo Almeida, Carroll~L. Wainwright, Pamela Mishkin, Chong Zhang, Sandhini Agarwal, Katarina Slama, Alex Ray, John Schulman, Jacob Hilton, Fraser Kelton, Luke Miller, Maddie Simens, Amanda Askell, Peter Welinder, Paul Christiano, Jan Leike, and Ryan Lowe. 2022.
\newblock \href {http://arxiv.org/abs/2203.02155} {Training language models to follow instructions with human feedback}.

\bibitem[{Pan and Zeng(2023)}]{pan2023llms}
Keyu Pan and Yawen Zeng. 2023.
\newblock \href {http://arxiv.org/abs/2307.16180} {Do llms possess a personality? making the mbti test an amazing evaluation for large language models}.

\bibitem[{Penedo et~al.(2023)Penedo, Malartic, Hesslow, Cojocaru, Cappelli, Alobeidli, Pannier, Almazrouei, and Launay}]{refinedweb}
Guilherme Penedo, Quentin Malartic, Daniel Hesslow, Ruxandra Cojocaru, Alessandro Cappelli, Hamza Alobeidli, Baptiste Pannier, Ebtesam Almazrouei, and Julien Launay. 2023.
\newblock \href {http://arxiv.org/abs/2306.01116} {The {R}efined{W}eb dataset for {F}alcon {LLM}: outperforming curated corpora with web data, and web data only}.
\newblock \emph{arXiv preprint arXiv:2306.01116}.

\bibitem[{Pittenger(1993)}]{e73f6545-174d-366a-9e0e-d380358fcd01}
David~J. Pittenger. 1993.
\newblock \href {http://www.jstor.org/stable/1170497} {The utility of the myers-briggs type indicator}.
\newblock \emph{Review of Educational Research}, 63(4):467--488.

\bibitem[{Radford et~al.(2019)Radford, Wu, Child, Luan, Amodei, and Sutskever}]{radford2019language}
Alec Radford, Jeff Wu, Rewon Child, David Luan, Dario Amodei, and Ilya Sutskever. 2019.
\newblock \href {https://api.semanticscholar.org/CorpusID:160025533} {Language models are unsupervised multitask learners}.

\bibitem[{{Serapio-Garc{\'\i}a} et~al.(2023){Serapio-Garc{\'\i}a}, Safdari, Crepy, Sun, Fitz, Romero, Abdulhai, Faust, and Matari{\'c}}]{serapio-garciaPersonalityTraitsLarge2023a}
Greg {Serapio-Garc{\'\i}a}, Mustafa Safdari, Cl{\'e}ment Crepy, Luning Sun, Stephen Fitz, Peter Romero, Marwa Abdulhai, Aleksandra Faust, and Maja Matari{\'c}. 2023.
\newblock \href {http://arxiv.org/abs/2307.00184} {Personality {{Traits}} in {{Large Language Models}}}.

\bibitem[{Srivastava et~al.(2014)Srivastava, Hinton, Krizhevsky, Sutskever, and Salakhutdinov}]{JMLR:v15:srivastava14a}
Nitish Srivastava, Geoffrey Hinton, Alex Krizhevsky, Ilya Sutskever, and Ruslan Salakhutdinov. 2014.
\newblock \href {http://jmlr.org/papers/v15/srivastava14a.html} {Dropout: A simple way to prevent neural networks from overfitting}.
\newblock \emph{Journal of Machine Learning Research}, 15(56):1929--1958.

\bibitem[{Stajner and Yenikent(2020)}]{stajner-yenikent-2020-survey}
Sanja Stajner and Seren Yenikent. 2020.
\newblock \href {https://doi.org/10.18653/v1/2020.coling-main.553} {A survey of automatic personality detection from texts}.
\newblock In \emph{Proceedings of the 28th International Conference on Computational Linguistics}, pages 6284--6295, Barcelona, Spain (Online). International Committee on Computational Linguistics.

\bibitem[{Storey(2018)}]{dylan_storey_2018_1482951}
Dylan Storey. 2018.
\newblock \href {https://doi.org/10.5281/zenodo.1482951} {Myers briggs personality tags on reddit data}.

\bibitem[{Suman et~al.(2020)Suman, Gupta, Saha, and Bhattacharyya}]{suman-etal-2020-multi}
Chanchal Suman, Aditya Gupta, Sriparna Saha, and Pushpak Bhattacharyya. 2020.
\newblock \href {https://aclanthology.org/2020.icon-main.42} {A multi-modal personality prediction system}.
\newblock In \emph{Proceedings of the 17th International Conference on Natural Language Processing (ICON)}, pages 317--322, Indian Institute of Technology Patna, Patna, India. NLP Association of India (NLPAI).

\bibitem[{Taori et~al.(2023)Taori, Gulrajani, Zhang, Dubois, Li, Guestrin, Liang, and Hashimoto}]{alpaca}
Rohan Taori, Ishaan Gulrajani, Tianyi Zhang, Yann Dubois, Xuechen Li, Carlos Guestrin, Percy Liang, and Tatsunori~B. Hashimoto. 2023.
\newblock Stanford alpaca: An instruction-following llama model.
\newblock \url{https://github.com/tatsu-lab/stanford_alpaca}.

\bibitem[{Touvron et~al.(2023)Touvron, Martin, Stone, Albert, Almahairi, Babaei, Bashlykov, Batra, Bhargava, Bhosale, Bikel, Blecher, Ferrer, Chen, Cucurull, Esiobu, Fernandes, Fu, Fu, Fuller, Gao, Goswami, Goyal, Hartshorn, Hosseini, Hou, Inan, Kardas, Kerkez, Khabsa, Kloumann, Korenev, Koura, Lachaux, Lavril, Lee, Liskovich, Lu, Mao, Martinet, Mihaylov, Mishra, Molybog, Nie, Poulton, Reizenstein, Rungta, Saladi, Schelten, Silva, Smith, Subramanian, Tan, Tang, Taylor, Williams, Kuan, Xu, Yan, Zarov, Zhang, Fan, Kambadur, Narang, Rodriguez, Stojnic, Edunov, and Scialom}]{touvron2023llama}
Hugo Touvron, Louis Martin, Kevin Stone, Peter Albert, Amjad Almahairi, Yasmine Babaei, Nikolay Bashlykov, Soumya Batra, Prajjwal Bhargava, Shruti Bhosale, Dan Bikel, Lukas Blecher, Cristian~Canton Ferrer, Moya Chen, Guillem Cucurull, David Esiobu, Jude Fernandes, Jeremy Fu, Wenyin Fu, Brian Fuller, Cynthia Gao, Vedanuj Goswami, Naman Goyal, Anthony Hartshorn, Saghar Hosseini, Rui Hou, Hakan Inan, Marcin Kardas, Viktor Kerkez, Madian Khabsa, Isabel Kloumann, Artem Korenev, Punit~Singh Koura, Marie-Anne Lachaux, Thibaut Lavril, Jenya Lee, Diana Liskovich, Yinghai Lu, Yuning Mao, Xavier Martinet, Todor Mihaylov, Pushkar Mishra, Igor Molybog, Yixin Nie, Andrew Poulton, Jeremy Reizenstein, Rashi Rungta, Kalyan Saladi, Alan Schelten, Ruan Silva, Eric~Michael Smith, Ranjan Subramanian, Xiaoqing~Ellen Tan, Binh Tang, Ross Taylor, Adina Williams, Jian~Xiang Kuan, Puxin Xu, Zheng Yan, Iliyan Zarov, Yuchen Zhang, Angela Fan, Melanie Kambadur, Sharan Narang, Aurelien Rodriguez, Robert Stojnic, Sergey Edunov, and Thomas
  Scialom. 2023.
\newblock \href {http://arxiv.org/abs/2307.09288} {Llama 2: Open foundation and fine-tuned chat models}.

\bibitem[{tse Huang et~al.(2023)tse Huang, Wang, Lam, Li, Jiao, and Lyu}]{huang2023revisiting}
Jen tse Huang, Wenxuan Wang, Man~Ho Lam, Eric~John Li, Wenxiang Jiao, and Michael~R. Lyu. 2023.
\newblock \href {http://arxiv.org/abs/2305.19926} {Revisiting the reliability of psychological scales on large language models}.

\bibitem[{V~Ganesan et~al.(2023)V~Ganesan, Lal, Nilsson, and Schwartz}]{v-ganesan-etal-2023-systematic}
Adithya V~Ganesan, Yash~Kumar Lal, August Nilsson, and H.~Schwartz. 2023.
\newblock \href {https://doi.org/10.18653/v1/2023.wassa-1.34} {Systematic evaluation of {GPT}-3 for zero-shot personality estimation}.
\newblock In \emph{Proceedings of the 13th Workshop on Computational Approaches to Subjectivity, Sentiment, {\&} Social Media Analysis}, pages 390--400, Toronto, Canada. Association for Computational Linguistics.

\bibitem[{Vu et~al.(2018)Vu, Flekova, Jiang, and Gurevych}]{vu-etal-2018-lexical}
Xuan-Son Vu, Lucie Flekova, Lili Jiang, and Iryna Gurevych. 2018.
\newblock \href {https://aclanthology.org/2018.gwc-1.20} {Lexical-semantic resources: yet powerful resources for automatic personality classification}.
\newblock In \emph{Proceedings of the 9th Global Wordnet Conference}, pages 172--181, Nanyang Technological University (NTU), Singapore. Global Wordnet Association.

\bibitem[{Wang et~al.(2023)Wang, Kordi, Mishra, Liu, Smith, Khashabi, and Hajishirzi}]{wang-etal-2023-self-instruct}
Yizhong Wang, Yeganeh Kordi, Swaroop Mishra, Alisa Liu, Noah~A. Smith, Daniel Khashabi, and Hannaneh Hajishirzi. 2023.
\newblock \href {https://doi.org/10.18653/v1/2023.acl-long.754} {Self-instruct: Aligning language models with self-generated instructions}.
\newblock In \emph{Proceedings of the 61st Annual Meeting of the Association for Computational Linguistics (Volume 1: Long Papers)}, pages 13484--13508, Toronto, Canada. Association for Computational Linguistics.

\bibitem[{Wei et~al.(2022{\natexlab{a}})Wei, Bosma, Zhao, Guu, Yu, Lester, Du, Dai, and Le}]{wei2022finetuned}
Jason Wei, Maarten Bosma, Vincent~Y. Zhao, Kelvin Guu, Adams~Wei Yu, Brian Lester, Nan Du, Andrew~M. Dai, and Quoc~V. Le. 2022{\natexlab{a}}.
\newblock \href {http://arxiv.org/abs/2109.01652} {Finetuned language models are zero-shot learners}.

\bibitem[{Wei et~al.(2022{\natexlab{b}})Wei, Tay, Bommasani, Raffel, Zoph, Borgeaud, Yogatama, Bosma, Zhou, Metzler, Chi, Hashimoto, Vinyals, Liang, Dean, and Fedus}]{wei2022emergent}
Jason Wei, Yi~Tay, Rishi Bommasani, Colin Raffel, Barret Zoph, Sebastian Borgeaud, Dani Yogatama, Maarten Bosma, Denny Zhou, Donald Metzler, Ed~H. Chi, Tatsunori Hashimoto, Oriol Vinyals, Percy Liang, Jeff Dean, and William Fedus. 2022{\natexlab{b}}.
\newblock \href {https://openreview.net/forum?id=yzkSU5zdwD} {Emergent abilities of large language models}.
\newblock \emph{Transactions on Machine Learning Research}.
\newblock Survey Certification.

\bibitem[{Wei et~al.(2022{\natexlab{c}})Wei, Wang, Schuurmans, Bosma, ichter, Xia, Chi, Le, and Zhou}]{NEURIPS2022_9d560961}
Jason Wei, Xuezhi Wang, Dale Schuurmans, Maarten Bosma, brian ichter, Fei Xia, Ed~Chi, Quoc~V Le, and Denny Zhou. 2022{\natexlab{c}}.
\newblock \href {https://proceedings.neurips.cc/paper_files/paper/2022/file/9d5609613524ecf4f15af0f7b31abca4-Paper-Conference.pdf} {Chain-of-thought prompting elicits reasoning in large language models}.
\newblock In \emph{Advances in Neural Information Processing Systems}, volume~35, pages 24824--24837. Curran Associates, Inc.

\bibitem[{Wen et~al.(2023)Wen, Ke, Sun, Zhang, Li, Bai, and Huang}]{wen2023unveiling}
Jiaxin Wen, Pei Ke, Hao Sun, Zhexin Zhang, Chengfei Li, Jinfeng Bai, and Minlie Huang. 2023.
\newblock \href {https://openreview.net/forum?id=u69aCtohTC} {Unveiling the implicit toxicity in large language models}.
\newblock In \emph{The 2023 Conference on Empirical Methods in Natural Language Processing}.

\bibitem[{Yao et~al.(2023)Yao, Aminabadi, Ruwase, Rajbhandari, Wu, Awan, Rasley, Zhang, Li, Holmes, Zhou, Wyatt, Smith, Kurilenko, Qin, Tanaka, Che, Song, and He}]{yao2023dschat}
Zhewei Yao, Reza~Yazdani Aminabadi, Olatunji Ruwase, Samyam Rajbhandari, Xiaoxia Wu, Ammar~Ahmad Awan, Jeff Rasley, Minjia Zhang, Conglong Li, Connor Holmes, Zhongzhu Zhou, Michael Wyatt, Molly Smith, Lev Kurilenko, Heyang Qin, Masahiro Tanaka, Shuai Che, Shuaiwen~Leon Song, and Yuxiong He. 2023.
\newblock \href {http://arxiv.org/abs/2308.01320} {Deepspeed-chat: Easy, fast and affordable rlhf training of chatgpt-like models at all scales}.

\bibitem[{Zeng et~al.(2023)Zeng, Liu, Du, Wang, Lai, Ding, Yang, Xu, Zheng, Xia, Tam, Ma, Xue, Zhai, Chen, Zhang, Dong, and Tang}]{zeng2022glm}
Aohan Zeng, Xiao Liu, Zhengxiao Du, Zihan Wang, Hanyu Lai, Ming Ding, Zhuoyi Yang, Yifan Xu, Wendi Zheng, Xiao Xia, Weng~Lam Tam, Zixuan Ma, Yufei Xue, Jidong Zhai, Wenguang Chen, Peng Zhang, Yuxiao Dong, and Jie Tang. 2023.
\newblock \href {http://arxiv.org/abs/2210.02414} {Glm-130b: An open bilingual pre-trained model}.

\bibitem[{Zhang et~al.(2024)Zhang, Dong, Li, Zhang, Sun, Wang, Li, Hu, Zhang, Wu, and Wang}]{zhang2024instruction}
Shengyu Zhang, Linfeng Dong, Xiaoya Li, Sen Zhang, Xiaofei Sun, Shuhe Wang, Jiwei Li, Runyi Hu, Tianwei Zhang, Fei Wu, and Guoyin Wang. 2024.
\newblock \href {http://arxiv.org/abs/2308.10792} {Instruction tuning for large language models: A survey}.

\bibitem[{Zhou et~al.(2023{\natexlab{a}})Zhou, Liu, Xu, Iyer, Sun, Mao, Ma, Efrat, Yu, Yu, Zhang, Ghosh, Lewis, Zettlemoyer, and Levy}]{zhou2023lima}
Chunting Zhou, Pengfei Liu, Puxin Xu, Srini Iyer, Jiao Sun, Yuning Mao, Xuezhe Ma, Avia Efrat, Ping Yu, Lili Yu, Susan Zhang, Gargi Ghosh, Mike Lewis, Luke Zettlemoyer, and Omer Levy. 2023{\natexlab{a}}.
\newblock \href {http://arxiv.org/abs/2305.11206} {Lima: Less is more for alignment}.

\bibitem[{Zhou et~al.(2023{\natexlab{b}})Zhou, Sch{\"a}rli, Hou, Wei, Scales, Wang, Schuurmans, Cui, Bousquet, Le, and Chi}]{zhou2023leasttomost}
Denny Zhou, Nathanael Sch{\"a}rli, Le~Hou, Jason Wei, Nathan Scales, Xuezhi Wang, Dale Schuurmans, Claire Cui, Olivier Bousquet, Quoc~V Le, and Ed~H. Chi. 2023{\natexlab{b}}.
\newblock \href {https://openreview.net/forum?id=WZH7099tgfM} {Least-to-most prompting enables complex reasoning in large language models}.
\newblock In \emph{The Eleventh International Conference on Learning Representations}.

\bibitem[{Ziegler et~al.(2020)Ziegler, Stiennon, Wu, Brown, Radford, Amodei, Christiano, and Irving}]{ziegler2020finetuning}
Daniel~M. Ziegler, Nisan Stiennon, Jeffrey Wu, Tom~B. Brown, Alec Radford, Dario Amodei, Paul Christiano, and Geoffrey Irving. 2020.
\newblock \href {http://arxiv.org/abs/1909.08593} {Fine-tuning language models from human preferences}.

\end{thebibliography}
\bibliographystyle{acl_natbib}

\appendix

% TODO: 补全所有的附录

\section{MBTI Personality Assessment}
\label{sec:mbti_personality_assessment}

We collected publicly available MBTI questionnaires and refined them into a comprehensive 200-item MBTI Assessment (50 items per dichotomous dimension)~\cite{pan2023llms}\footnote{\label{fn:url1}\url{https://www.16personalities.com/}}\footnote{\label{fn:url2}\url{https://www.truity.com/}}\footnote{\label{fn:url3}\url{https://www.humanmetrics.com/}}. As shown in Table \ref{item_distribution}, the Attitude dimension, encompassing both Extroversion and Introversion traits, is evaluated with 50 items. The other three dimensions follow the same pattern.

\begin{table}[htb]
    \centering
    \resizebox{0.35\textwidth}{!}{
    \begin{tabular}{cccccccccccccccccccccccc}
    \toprule
    \textbf{Traits} & \textbf{Items}\\
    \midrule
    Extroversion/Introversion & 50 \\
    Sensing/Intuition & 50 \\
    Thinking/Feeling & 50 \\
    Judging/Perceiving & 50 \\
    \bottomrule
    \end{tabular}
    }
    \caption{Item Distribution.}
    \label{item_distribution}
\end{table}

Some item examples are shown in Table \ref{item_examples}.

\begin{table}[htb]
    \centering
    %\resizebox{0.48\textwidth}{!}{
    \small
    \begin{tabularx}{0.48\textwidth}{X}
    %\begin{tabular}{cccccccccccccccccccccccc}
    \toprule
    \textbf{Item Example}\\
    \midrule
    You enjoy having a wide social circle. \\
    Option A: Yes. \\
    Option B: No. You prefer to be left alone if you have a choice. \\
    \midrule
    You dislike unexpected occurrences, which disrupt your plans. \\
    Option A: Yes. \\
    Option B: No. \\
    \midrule
    People who know you tend to describe you as \\
    Option A: Logical and Clarity. \\
    Option B: Passionate and Sensitive. \\
    \bottomrule
    %\end{tabular}
    %}
    \end{tabularx}
    \caption{Item Examples.}
    \label{item_examples}
\end{table}

% TODO 
%\section{Prompt for Data Generation}
%\label{sec:template_data_generation}

%Utilizing the Least-to-Most~\cite{zhou2023leasttomost} mindset, we partitioned the data generation process into two stages: initially crafting questions rooted in Opposite Trait Description, followed by eliciting responses from Prompt-induced LLMs. In this section, we present some examples of question generation prompt and response generation prompt. As shown in Table \ref{question_prompt_template}, to aid the model in discerning between two opposite traits, we integrated descriptions of two opposite traits from the same dimension into the prompt. Moreover, as depicted in Table \ref{response_prompt_template}, we induced models to embody specific personality traits in generating responses to corresponding questions.

\section{Answer Extractor}
\label{sec:appendix_answer_extractor}

\begin{table*}[htbp]
    \centering
    \resizebox{\textwidth}{!}{
    \begin{tabular}{p{15cm}}
    \toprule
    \textbf{Pretrain Data Examples}\\
    \midrule
    You are totally replying to the wrong person. If you want to argue with this moron about his misguided thoughts on politics please copy and paste your reply to HIS post, not the one where I point out how stupid his post is.|||1982-2004. All generations are about 20 years, give or take 5-ish. About the length of a phase of life.|||Comment seemed cooler when I misread your name as FARGO.|||Per how I adjusted his recipe it's putting me at 1555 calories, or at least that's what the program is saying? Am I calculating something wrong?|||I agree, but the whole "not change anything" line doesn't exactly say that he will make changes to how they currently use the fund.|||I'm hanging out for some 30s musical classics!|||Supported. I don't think it's gonna make it...|||\\
    \bottomrule
    \end{tabular}}
    \caption{Pretrain Data Examples.}
    \label{dataset_format_pretrain}
\end{table*}

\begin{table}[htb]
    \centering
    \resizebox{0.5\textwidth}{!}{
    \begin{tabular}{cccccccccccccccccccccccc}
    \toprule
    \textbf{Dataset} & \textbf{Precision} & \textbf{Recall} & \textbf{Macro-F1} & \textbf{Accuracy} \\
    \midrule
    valid & $95.47\%$ & $93.94\%$ & $94.65\%$ & $95.95\%$\\
    \bottomrule
    \end{tabular}
    }
    \caption{Answer Extractor Performance}
    \label{answer_extractor_performance}
\end{table}

Recognizing the open-ended nature of LLMs~\cite{wen2023unveiling}, 
LLMs may not always response with an answer directly. Thus, we trained an Answer Extractor to identify numerical information in model responses. For this purpose, we labeled 3774 samples, randomly splitting 420 samples for validation and tuned falcon-7B-instruct~\cite{almazrouei2023falcon, refinedweb} as the Answer Extractor. 

As shown in Table \ref{answer_extractor_performance}, the answer extractor achieved precision, recall, f1, and accuracy scores well above $90\%$ on the test set.

\section{Datasets}
\label{sec:training_data}

% TODO: 详细说明
In the main body, we elaborated on our data generation methodology for each training method. In this section, we present additional details of training datasets, including the prompts we used, examples of training instances for each training stage and the summary statistics.

\noindent \textbf{Continual Pretrain.} We amalgamated and refined existing datasets annotated with human personality predictions\footnote{\url{https://www.kaggle.com/datasets/zeyadkhalid/mbti-personality-types-500-dataset}}\footnote{\url{https://www.kaggle.com/datasets/mazlumi/mbti-personality-type-twitter-dataset}}\footnote{\url{https://www.kaggle.com/datasets/datasnaek/mbti-type}}\footnote{\url{https://zenodo.org/records/1482951}}. The data distribution can be referred to Figure \ref{fig:pretrain_data}. And the pre-training corpus format is detailed in Table \ref{dataset_format_pretrain}. Posts from each personality are delimited by `|||'. Noise is evident within the corpus.

\noindent \textbf{SFT.} Utilizing mentioned before, we partitioned the data generation process into two stages: initially crafting questions rooted in Opposite Trait Description, followed by eliciting responses from Prompt-induced LLMs. 

We present some examples of question generation prompt and response generation prompt. 
As shown in Table \ref{question_prompt_template}, to aid the model in discerning between two opposite traits, we integrated descriptions of two opposite traits from the same dimension into the prompt. 
Moreover, as depicted in Table \ref{response_prompt_template}, we induced models to embody specific personality traits in generating responses to corresponding questions. Additionally, in Table~\ref{sft_examples}, we present an example of generated SFT training data. The prompts utilized in our prompt induction process are delineated in Table \ref{role_play_prompt}.

\begin{table}[htb]
    \centering
    \small
    \begin{tabularx}{0.48\textwidth}{X}
    \toprule
    \textbf{SFT Instruction Pair Example}\\
    \midrule
    \{ \\
    \hangindent=1em \hspace*{1em}"Instruction input": \\
    \hangindent=2em \hspace*{2em}"Would you rather have a wide circle of acquaintances or a few deep, meaningful relationships?", \\
    \hangindent=1em \hspace*{1em}"Output": \\
    \hangindent=2em \hspace*{2em}"I would definitely prefer to have a wide circle of acquaintances! I love meeting new people and building connections with as many individuals as possible. I thrive on interacting with others and getting to know a diverse range of personalities. It's invigorating for me to have a wide network of friends and acquaintances to socialize and engage with." \\
    \}\\
    \bottomrule
    \end{tabularx}
    \caption{SFT Instruction Pair Example.}
    \label{sft_examples}
\end{table}

\noindent \textbf{RLHF.} To control variables, we used questions from SFT instruction pairs for PPO training. Notably, Previous work has shown that reinforcement learning training process may cause models to suffer from performance decrease in language ability~\cite{ouyang2022training}. Therefore, we followed previous work~\cite{yao2023dschat, ouyang2022training} by incorporating autoregressive training into the PPO process with widely recognized Wikipedia datasets~\cite{wikidump}, ensuring the model retains its ability to generate responses smoothly. 

Wikipedia datasets we used are the widely recognized official datasets released on Hugging Face generally used for Language Modeling, which comprises pre-processed subsets in six languages. Each language's Wikipedia data are officially segmented into distinct segments. For instance, the English dataset consists of 6,458,670 samples, segmented into 41 segments. In our work, considering we do not need so many samples, thus we randomly used one English segment, which has 157,529 samples with an average word count of 1834.49 per sample.

For training reward model, we used prompt-induced LLMs to generate question-answer pairs, such as (Question, Extroversion-induced model response, Introversion-induced model response). In Table~\ref{rlhf_examples}, we present an example of the task input and output for training Extroversion reward model.

\begin{table}[htb]
    \centering
    \small
    \begin{tabularx}{0.48\textwidth}{X}
    \toprule
    \textbf{RLHF Reward Training Data Example}\\
    \midrule
    \{ \\
    \hangindent=1em \hspace*{1em}"Question": \\
    \hangindent=2em \hspace*{2em}"Would you rather have a wide circle of acquaintances or a few deep, meaningful relationships?", \\
    \hangindent=1em \hspace*{1em}"Extroversion-induced Model Response": \\
    \hangindent=2em \hspace*{2em}"I would definitely prefer to have a wide circle of acquaintances! I love meeting new people and building connections with as many individuals as possible. I thrive on interacting with others and getting to know a diverse range of personalities. It's invigorating for me to have a wide network of friends and acquaintances to socialize and engage with." \\
    \hangindent=1em \hspace*{1em}"Introversion-induced Model Response": \\
    \hangindent=2em \hspace*{2em}"I would definitely prefer a few deep, meaningful relationships over a wide circle of acquaintances. I feel more comfortable and content when I can invest my time and energy into nurturing close and intimate connections with a select few individuals. Large social gatherings or maintaining numerous superficial relationships tend to drain my energy, so I gravitate towards deeper, more meaningful interactions with a small group of trusted individuals." \\
    \}\\
    \bottomrule
    \end{tabularx}
    \caption{RLHF Reward Training Data Example.}
    \label{rlhf_examples}
\end{table}

Finally, as shown in Table~\ref{results:dataset_statistic}, we present the detailed summary statistics.

\begin{table*}[htbp]
      \centering
      \resizebox{\textwidth}{!}{%
      \small
      \begin{tabular}{cccccccccc}
        \toprule
        \textbf{Datasets} & \textbf{Total Tokens} & \textbf{Total Words} & \textbf{Total Sentences} & \textbf{Mean Tokens$_{T}$} & \textbf{Mean Words$_{T}$} & \textbf{Mean Sentences$_{T}$}   & \textbf{Mean Tokens$_{P}$} & \textbf{Mean Words$_{P}$} & \textbf{Mean Sentences$_{P}$} \\
        \midrule
        Continual Pre-train & 236119950 & 207619050 & 10588585 & 23611995.0 & 20761905.0 & 1058858.5 & 2951499.375 & 2595238.125 & 132357.3125 \\
        SFT & 20964546 & 21281067 & 1324143 & 291174.25 & 295570.375 & 18390.875 & 1164697.0 & 1182281.5 & 73563.5 \\
        RLHF-PPO & 5500422 & 5363298 & 180198 & 76394.75 & 74490.25 & 2502.75 & 305579.0 & 297961.0 & 10011.0 \\
        RLHF-Reward & 345321864 & 337057092 & 14992074 & 4796137.0 & 4681348.5 & 208223.25 & 19184548.0 & 18725394.0 & 832893.0 \\
        \bottomrule 
    \end{tabular}}
    \caption{Summary Statistics of Training Datasets. $T$ stands for trait data and $P$ stands for personality data.}
    \label{results:dataset_statistic}
\end{table*}

\section{Training}
\label{sec:rlhf_training}

\begin{figure}[htbp]
  \centering
  \includegraphics[width=0.48\textwidth]{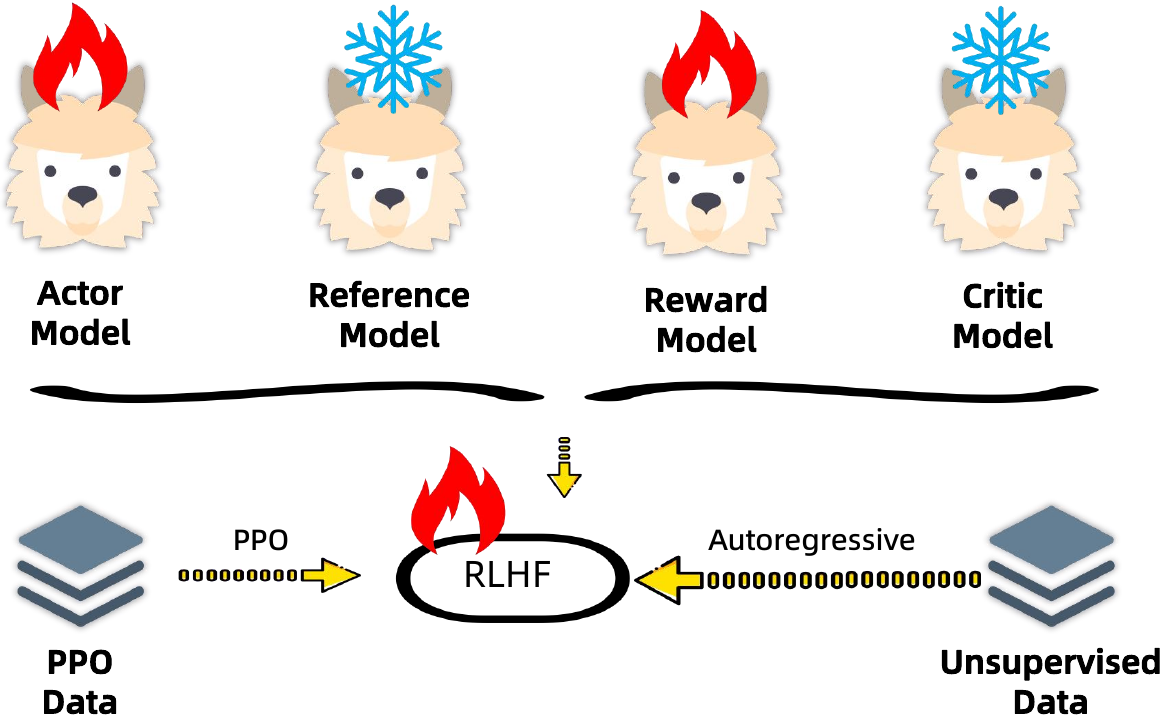}
  \caption{RLHF Training.}
  \label{fig:rlhf_training}
\end{figure}

\noindent \textbf{RLHF.} We adopted methodologies from InstructGPT~\cite{ouyang2022training} and DeepSpeed-Chat~\cite{yao2023dschat}, employing PPO-ptx~\cite{ouyang2022training} objective and Actor-Critic~\cite{NIPS1999_6449f44a} architecture. Figure \ref{fig:rlhf_training} illustrates the training process, where PPO-ptx introduces autoregressive objective during PPO training. As mentioned in main body and Appendix~\ref{sec:training_data}, we leveraged Wikipedia datasets as unsupervised training data. 

And each model is trained with its own reward model individually. For example, during the training of Llama2-chat-13B, we employed Llama2-chat-13B both as the Actor Model and the Reference Model. Additionally, we trained Llama2-chat-13B on the Reward Model training dataset, serving as both the Reward Model and the Critic Model.

We presented detailed performance of all reward models in Tables \ref{Llama_reward_model_performance}, \ref{Qwen_reward_model_performance} and \ref{ChatGLM2_reward_model_performance}. We observed high accuracy across all three models, with the performance of the reward model showing a strong correlation with model parameter size: Llama2-chat-13B > Qwen-chat-7B > ChatGLM2-6B. 
This may suggest a significant relationship between the performance of RLHF and the model size. 

%\section{Prompt Induction Prompts}
%\label{sec:appendix_role_play_prompt}

%The prompts utilized in our prompt induction process are delineated in Table \ref{role_play_prompt}.

\section{Scaling up Training Data for Continual Pre-train}
\label{sec:app_analyse}

\begin{figure}[htbp]
  \centering
  \includegraphics[width=0.48\textwidth, height=0.20\textheight]{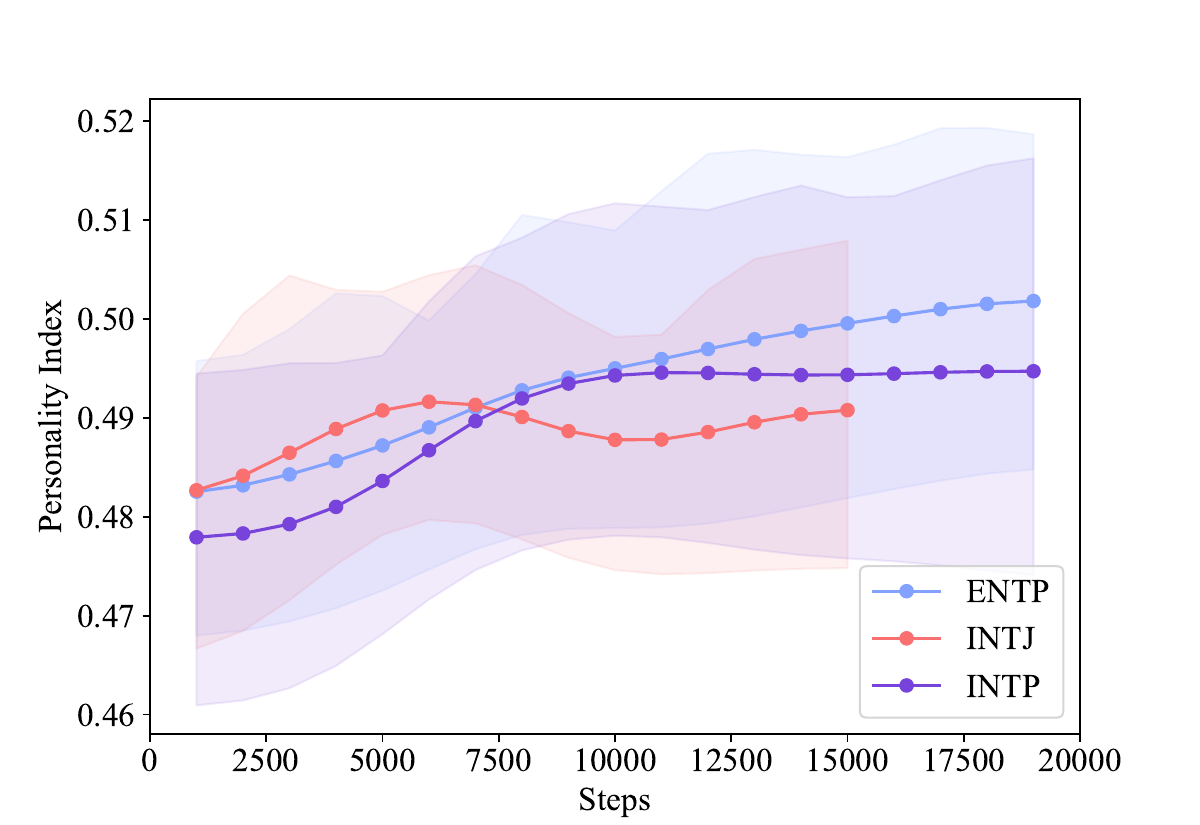}
  \caption{Continual Pre-train: Scaling up training data. Personality Index is the calculated mean of all trait proportions. A higher Personality Index indicates closer proximity to the specified personality.}
  \label{fig:large_pretrain_data}
\end{figure}

The minimal impact of continual pretrain control may be attributed to the more extensive dataset used during model pretraining, which inherently encompasses a mixed personality distribution. And the limited personality data fails to significantly influence its distribution. For additional validation, we enlarged the dataset size in specific personality control. We randomly selected three personalities and utilized all gathered samples for  training. 

As depicted in Figure \ref{fig:large_pretrain_data}, this led to a marginal improvement with increased data. This suggests that specific personality data can impact LLMs' synthetic personalities during pre-training and the control performance of Continual Pre-train is significantly influenced by the amount of personality data.

\begin{table*}[htbp]
    \centering
    \begin{tabular}{p{15cm}}
    \toprule
    \textbf{Question Generation Prompt Example} \\
    \midrule
    \textcolor{myblue}{Below, I need your help in generating 10 questions that can differentiate between the two personality traits of \texttt{Extraversion} \& \texttt{Introversion}.} \\\\

\textbf{Requirements}: \\
1.Questions should highlight the differences between the two personality traits of \texttt{Extraversion} \& \texttt{Introversion}. Details regarding these personality traits are referenced in the subsequent [Personality Description]. \\
2.Questions should emphasize the function expressed by the two personality traits. Refer to the following [Dimension Description]. \\
3.Please refrain from disclosing the content of [Personality Description] and [Dimension Description]. \\
4.Avoid generating duplicate questions. Any existing questions provided are listed in [Historical Questions]. \\\\

\textbf{[Dimension Description]} \\
\texttt{Extraversion} \& \texttt{Introversion} is about \texttt{**Orientation of Personal Energy**}: describes the way in which a person wants to interact with the world. \\\\

\textbf{[Personality Description]} \\
**Extraversion** refers to the act or state of being energized by the world outside the self. Extraverts enjoy socializing and tend to be more enthusiastic, assertive, talkative, and animated. They enjoy time spent with more people and find it less rewarding to spend time alone. They are Initiating, Expressive, Gregarious, Active and Enthusiastic. \\
Key characteristics: Directs energy outward. Gains energy from interaction. \\
**Introversion**, on the contrary, is the state of being predominately concerned with one’s inner world. Introverts prefer self-reflection to social interactions. They also prefer to observe before participating in an activity. Introverts tend to more quiet, ‘peaceful’, and reserved. Introverts *prefer* individual activities over social ones—this. They are Receiving, Contained, Intimate, Reflective and Quiet. \\
Key characteristics: Directs energy inward. Loses energy from interaction. \\\\

\textbf{[Historical Questions]} \\
None \\\\

\textcolor{myred}{Please generate 10 more questions below:}  \\
    \bottomrule
    \end{tabular}
    \caption{Question Generation Prompt. \textcolor{myblue}{Task Description}, Requirements, Dimension Description, Personality Description, Historical Questions, \textcolor{myred}{Task Instruction}.}
    \label{question_prompt_template}
\end{table*}

\begin{table*}[htbp]
    \centering
    \begin{tabular}{p{15cm}}
    \toprule
    \textbf{Response Generation Prompt Example} \\
    \midrule
    \textcolor{myblue}{Below, I need your help to embody a specified personality based on the given personality description and answer the corresponding questions:} \\\\

\textbf{[Dimension Description]} \\
\texttt{Extraversion} \& \texttt{Introversion} is about \texttt{**Orientation of Personal Energy**}: describes the way in which a person wants to interact with the world. \\\\

\textbf{[Personality Description]}\\
\texttt{**Extraversion**} refers to the act or state of being energized by the world outside the self. Extraverts enjoy socializing and tend to be more enthusiastic, assertive, talkative, and animated. They enjoy time spent with more people and find it less rewarding to spend time alone. They are Initiating, Expressive, Gregarious, Active and Enthusiastic. \\
Key characteristics: Directs energy outward. Gains energy from interaction. \\\\
 
\textbf{[Instruction]} \\
Now you need to embody a character with strong \texttt{**Extraversion**}(E) trait based on the given personality description. \\
Please answer from a first-person perspective.
Please try not to use overly absolute and unnatural words, like "definitely", "absolutely" and so on. \\\\

\textbf{[Question]} \\
When making plans, do you tend to seek out group activities or prefer solo pursuits? \\\\

\textbf{[Answer]} \\
    \bottomrule
    \end{tabular}
    \caption{Response Generation Prompt. \textcolor{myblue}{Task Description}, Dimension Description, Personality Description, Instruction, Question, Answer Flag.}
    \label{response_prompt_template}
\end{table*}

\begin{table*}[htbp]
    \centering
    \begin{tabular}{p{15cm}}
    \toprule
    \textbf{Specific Trait Role-Play Prompt Example - Extroversion} \\
    \midrule
    \textcolor{myblue}{Please embody the designated persona according to the provided personality description and answer the following questions imitating the specified persona:} \\\\
    
    \textbf{Personality Description}: \\
    \texttt{**Extraversion**} refers to the act or state of being energized by the world outside the self. Extraverts enjoy socializing and tend to be more enthusiastic, assertive, talkative, and animated. They enjoy time spent with more people and find it less rewarding to spend time alone. They are Initiating, Expressive, Gregarious, Active and Enthusiastic. \\\\
    
    \textbf{Instructions}: \\
    \textcolor{myred}{Below, please engage in role-playing based on the given personality description and portray a persona. A role with Extroverted(E) trait.} \\\\
    
    \midrule
    \textbf{Specific Personality Role-Play Prompt Example - ENTJ} \\
    \midrule 
    \textcolor{myblue}{Here is a role-playing task where you are required to assume a designated persona as described and answer the related questions:}\\\\
    
    \textbf{Personality Description}:\\
    \texttt{**Extraversion**} \\
    \texttt{**Extraversion**} refers to the act or state of being energized by the world outside the self. Extraverts enjoy socializing and tend to be more enthusiastic, assertive, talkative, and animated. They enjoy time spent with more people and find it less rewarding to spend time alone. They are Initiating, Expressive, Gregarious, Active and Enthusiastic.\\
    \texttt{**Intuition**}\\
    \texttt{**Intuition**} refers to how people process data. Intuitive people are keener to the meaning and patterns behind information. Intuitive people are more focused on how the present would affect the future. They are readily able to grasp different possibilities and abstract concepts. They easily see the big picture rather than the details. They are Abstract, Imaginative, Conceptual, Theoretical and Original.\\
    \texttt{**Feeling**}\\
    \texttt{**Feeling**} people are more subjective. They base their decisions on principles and personal values. When making decisions, they consider other people’s feelings and take it in account. It is in their best mind to maintain harmony among a group. They are more governed by their heart. They are Empathetic, Compassionate, Accommodating, Accepting and Tender.\\
    \texttt{**Judging**}\\
    \texttt{**Judging**} refers to how people outwardly display themselves when making decisions. Judging people have a tendency to be organized and prompt. They like order prefer outlined schedules to working extemporaneously. They prefer plans. They find the outcome more rewarding than the process of creating something. Judging people seek closure. They are Systematic, Planful, Early Starting, Scheduled and Methodical.\\\\
    
    \textbf{Instructions}:\\
    \textcolor{myred}{Right now, you need to embody a persona based on the provided personality description.A role with Extroverted Intuition Feeling Judging(ENFJ) personality.}  \\
    \bottomrule
    \end{tabular}
    \caption{Role-Play Prompt Examples. \textcolor{myblue}{Task Description}, Personality Description, \textcolor{myred}{Task Instruction}. For each prompt component, we constructed five utterances with identical semantics but different textual forms.}
    \label{role_play_prompt}
\end{table*}

%The primary performance gap between RLHF-control and SFT-control stems from RLHF-control's weaker performance on the 6-7B small model(Qwen-chat-7B and ChatGLM2-6B), while SFT-control shows better generalizability. We suggest that this difference may arise from both the reward model and actor model experiencing performance decline due to reduced parameter size.

\begin{table*}[htb]
    \centering
    \resizebox{\textwidth}{!}{
    \begin{tabular}{cccccccccccccccccccccccc}
    \toprule
    \textbf{Model} & \textbf{Control} & \textbf{Accuracy}($\uparrow$) & \textbf{Chosen Score}($\uparrow$) & \textbf{Rejected Score}($\downarrow$) & \textbf{Diff}($\uparrow$) \\
    \midrule
    \multirow{24}{*}{\textbf{Llama2-chat-13B}} &E& 99.40\% & 19.14& -12.93& 32.07\\
&I& 100.00\% & 23.89& -21.61& 45.50\\
&S& 99.75\% & 19.34& -25.10& 44.44\\
&N& 99.85\% & 22.39& -30.07& 52.46\\
&T& 99.75\% & 15.72& -16.76& 32.48\\
&F& 100.00\% & 6.70& -26.09& 32.79\\
&J& 99.85\% & 10.44& -13.53& 23.97\\
&P& 100.00\% & 27.76& -21.13& 48.89\\
&ENFJ& 99.71\% & 17.57& -30.09& 47.67\\
&ENFP& 99.88\% & 27.32& -28.22& 55.53\\
&ENTJ& 99.81\% & 16.96& -29.84& 46.80\\
&ENTP& 99.85\% & 27.95& -23.90& 51.85\\
&ESFJ& 99.84\% & 20.07& -22.83& 42.90\\
&ESFP& 99.90\% & 26.27& -21.26& 47.53\\
&ESTJ& 99.88\% & 32.13& -32.86& 64.99\\
&ESTP& 99.84\% & 25.97& -28.59& 54.56\\
&INFJ& 99.86\% & 18.25& -31.53& 49.78\\
&INFP& 99.94\% & 29.66& -30.97& 60.63\\
&INTJ& 99.94\% & 35.02& -29.60& 64.62\\
&INTP& 99.76\% & 16.26& -38.13& 54.40\\
&ISFJ& 99.81\% & 20.23& -28.75& 48.98\\
&ISFP& 99.90\% & 28.14& -28.50& 56.64\\
&ISTJ& 99.91\% & 27.41& -44.64& 72.05\\
&ISTP& 99.83\% & 27.27& -34.86& 62.13\\
    \midrule
    Mean Score & & 99.84\% & 22.58& -27.16& 49.74\\
    \bottomrule
    \end{tabular}
    }
    \caption{Llama2-chat-13B Reward Model Performance}
    \label{Llama_reward_model_performance}
\end{table*}

\begin{table*}[htb]
    \centering
    \resizebox{\textwidth}{!}{
    \begin{tabular}{cccccccccccccccccccccccc}
    \toprule
    \textbf{Model} & \textbf{Control} & \textbf{Accuracy}($\uparrow$) & \textbf{Chosen Score}($\uparrow$) & \textbf{Rejected Score}($\downarrow$) & \textbf{Diff}($\uparrow$) \\
    \midrule
    \multirow{24}{*}{\textbf{Qwen-chat-7B}} &E& 99.45\% & 16.13& -3.87& 20.00\\
&I& 99.85\% & 15.53& 1.43& 14.09\\
&S& 99.75\% & 12.13& -0.28& 12.41\\
&N& 99.85\% & 17.21& 4.68& 12.53\\
&T& 99.30\% & 10.71& 3.88& 6.84\\
&F& 99.90\% & 7.38& -9.96& 17.34\\
&J& 99.70\% & 12.04& 4.07& 7.97\\
&P& 100.00\% & 20.00& -1.82& 21.83\\
&ENFJ& 99.73\% & 14.76& -1.84& 16.60\\
&ENFP& 99.84\% & 14.85& -6.53& 21.37\\
&ENTJ& 99.79\% & 14.90& -3.25& 18.15\\
&ENTP& 99.81\% & 14.71& -5.02& 19.72\\
&ESFJ& 99.64\% & 15.26& -0.60& 15.87\\
&ESFP& 99.76\% & 13.23& -3.81& 17.04\\
&ESTJ& 99.78\% & 16.53& -3.47& 20.00\\
&ESTP& 99.76\% & 16.61& -1.07& 17.68\\
&INFJ& 99.75\% & 15.87& 0.15& 15.73\\
&INFP& 99.84\% & 15.42& -2.80& 18.22\\
&INTJ& 99.88\% & 15.84& -6.04& 21.87\\
&INTP& 99.81\% & 15.70& -2.67& 18.37\\
&ISFJ& 99.65\% & 16.20& 1.48& 14.72\\
&ISFP& 99.85\% & 15.07& -4.16& 19.23\\
&ISTJ& 99.93\% & 16.39& -7.23& 23.62\\
&ISTP& 99.74\% & 19.41& -0.20& 19.61\\
    \midrule
    Mean Score & & 99.76\% & 15.08& -2.04& 17.12\\
    \bottomrule
    \end{tabular}
    }
    \caption{Qwen-chat-7B Reward Model Performance}
    \label{Qwen_reward_model_performance}
\end{table*}

\begin{table*}[htb]
    \centering
    \resizebox{\textwidth}{!}{
    \begin{tabular}{cccccccccccccccccccccccc}
    \toprule
    \textbf{Model} & \textbf{Control} & \textbf{Accuracy}($\uparrow$) & \textbf{Chosen Score}($\uparrow$) & \textbf{Rejected Score}($\downarrow$) & \textbf{Diff}($\uparrow$) \\
    \midrule
    \multirow{24}{*}{\textbf{ChatGLM2-6B}} &E& 98.85\% & 6.61& -2.95& 9.56\\
&I& 99.45\% & 8.17& -2.22& 10.38\\
&S& 99.70\% & 7.45& -4.37& 11.81\\
&N& 98.90\% & 7.24& -1.80& 9.04\\
&T& 97.20\% & 5.58& -0.28& 5.87\\
&F& 99.30\% & 6.63& -4.55& 11.19\\
&J& 98.80\% & 3.62& -4.47& 8.09\\
&P& 99.45\% & 9.23& -2.71& 11.94\\
&ENFJ& 98.89\% & 5.33& -6.77& 12.09\\
&ENFP& 99.53\% & 7.64& -3.92& 11.56\\
&ENTJ& 99.38\% & 6.17& -4.59& 10.76\\
&ENTP& 99.45\% & 7.47& -3.19& 10.65\\
&ESFJ& 98.96\% & 5.24& -7.22& 12.45\\
&ESFP& 99.09\% & 6.88& -6.72& 13.60\\
&ESTJ& 99.40\% & 7.28& -8.10& 15.38\\
&ESTP& 99.18\% & 6.06& -7.63& 13.69\\
&INFJ& 99.48\% & 6.27& -4.72& 11.00\\
&INFP& 99.70\% & 7.56& -4.11& 11.67\\
&INTJ& 99.73\% & 8.09& -4.67& 12.76\\
&INTP& 99.50\% & 6.56& -5.48& 12.04\\
&ISFJ& 99.40\% & 6.42& -4.24& 10.66\\
&ISFP& 99.61\% & 7.74& -5.18& 12.92\\
&ISTJ& 99.75\% & 8.43& -5.12& 13.55\\
&ISTP& 99.50\% & 7.03& -6.04& 13.07\\
    \midrule
    Mean Score & & 99.26\% & 6.86& -4.63& 11.49\\
    \bottomrule
    \end{tabular}
    }
    \caption{ChatGLM2-6B Reward Model Performance}
    \label{ChatGLM2_reward_model_performance}
\end{table*}

\end{document}